\documentclass{article}
\usepackage{booktabs}
\usepackage{amsmath,amssymb}
\usepackage{algpseudocode}
\usepackage{amssymb}
\usepackage{graphicx}
\usepackage{multirow}
\usepackage{lscape}
\usepackage{rotating}

\begin{document}
\title{Understanding Deep Learning Techniques for Image Segmentation}

\author{Swarnendu Ghosh, Nibaran Das, Ishita Das, Ujjwal Maulik}

\maketitle
\begin{abstract}
	The machine learning community has been overwhelmed by a plethora of deep learning based approaches. Many challenging computer vision tasks such as detection, localization, recognition and segmentation of objects in unconstrained environment are being efficiently addressed by various types of deep neural networks like convolutional neural networks, recurrent networks, adversarial networks, autoencoders and so on. While there have been plenty of analytical studies regarding the object detection or recognition domain, many new deep learning techniques have surfaced with respect to image segmentation techniques. This paper approaches these various deep learning techniques of image segmentation from an analytical perspective. The main goal of this work is to provide an intuitive understanding of the major techniques that has made significant contribution to the image segmentation domain. Starting from some of the traditional image segmentation approaches, the paper progresses describing the effect deep learning had on the image segmentation domain. Thereafter, most of the major segmentation algorithms have been logically categorized with paragraphs dedicated to their unique contribution. With an ample amount of intuitive explanations, the reader is expected to have an improved ability to visualize the internal dynamics of these processes. 
\end{abstract} 
\section{Introduction}
\begin{figure}[t]
	\includegraphics[width=0.45\textwidth]{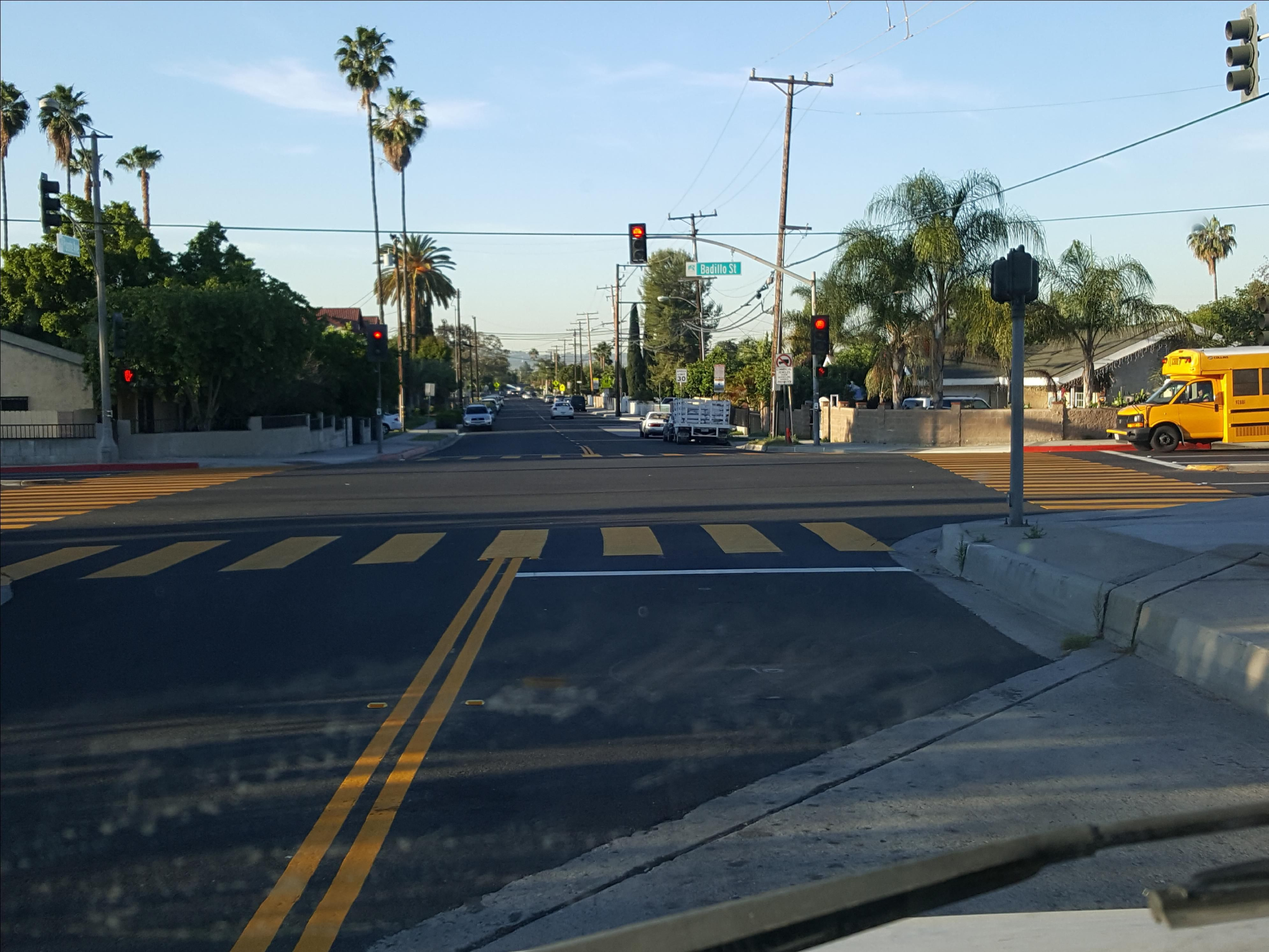}
	\includegraphics[width=0.45\textwidth]{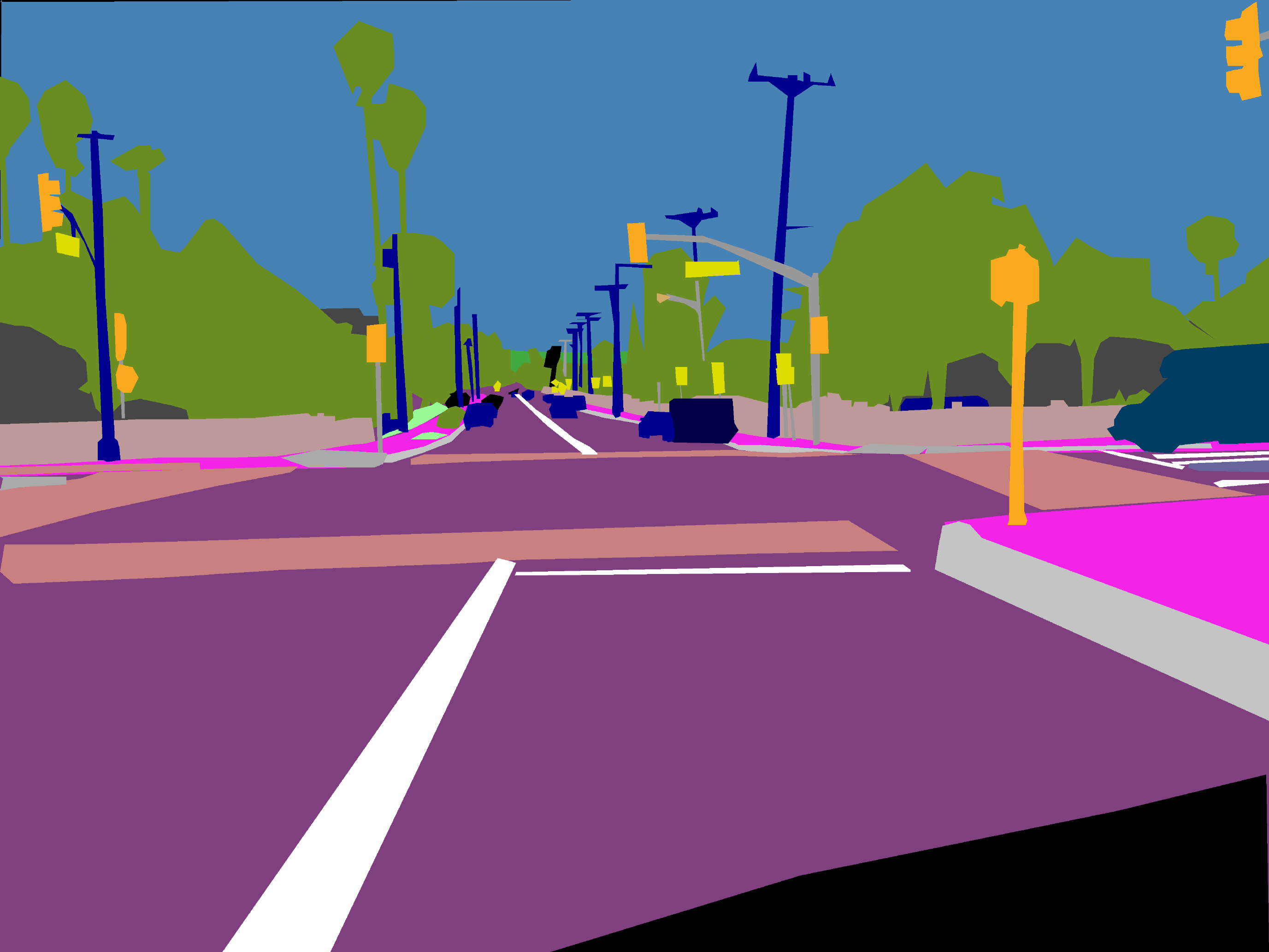}
	\caption{Semantic Image Segmentation(Samples from the Mapillary Vistas Dataset~\cite{data_mapillary})}
	\label{fig:imageseg}
\end{figure}

Image segmentation can be defined as a specific image processing technique which is used to divide an image into two or more meaningful regions. Image segmentation can also be seen as a process of defining boundaries between separate semantic entities in an image. From a more technical perspective, image segmentation is a process of assigning a label to each pixel in the image such that pixels with the same label are connected with respect to some visual or semantic property (Fig.~\ref{fig:imageseg}). 

Image segmentation subsumes a large class of finely related problems in computer vision. The most classic version is semantic segmentation ~\cite{survey6_deep}. In semantic segmentation, each pixel is classified into one of the predefined set of classes such that pixels belonging to the same class belongs to an unique semantic entity in the image. It is also worthy to note that the semantics in question depends not only on the data but also the problem that needs to be addressed. For example, for a pedestrian detection system, the whole body of person should belong to the same segment, however for a action recognition system, it might be necessary to segment different body parts into different classes. Other forms of image segmentation can focus on the most important object in a scene. A particular class of problem called saliency detection ~\cite{survey7_salient} is born from this. Other variants of this domain can be foreground background separation problems. In many systems like, image retrieval or visual question answering it is often necessary to count the number of objects. Instance specific segmentation addresses that issue. Instance specific segmentation is often coupled with object detection systems to detect and segment multiple instances of the same object\cite{dai_instance} in a scene. Segmentation in the temporal space is also a challenging domain and has various application. In object tracking scenarios, pixel level classification is not only performed in the spatial domain but also across time. Other applications in traffic analysis or surveillance needs to perform motion segmentation to analyze paths of moving objects. In the field of segmentation with lower semantic level, over-segmentation is also a common approach where images are divided into extremely small regions to ensure boundary adherence, at the cost of creating a lot of spurious edges. Over-segmentation algorithms are often combined with region merging techniques to perform image segmentation. Even simple color or texture segmentation also finds its use in various scenarios. Another important distinction between segmentation algorithms is the need of interactions from the user. While it is desirable to have fully automated systems, a little bit of interaction from the user can improve the quality of segmentation to a large extent. This is especially applicable where we are dealing with complex scenes or we do not posses an ample amount of data to train the system. 
\par Segmentation algorithms has several applications in the real world. In medical image processing ~\cite{survey5_medical} as well we need to localize various abnormalities like aneurysms~\cite{app1_aneurysm}, tumors~\cite{app2_tumor}, cancerous elements like melanoma detection~\cite{app3_melanoma}, or specific organs during surgeries~\cite{app4_surgery}. Another domain where segmentation is important is surveillance. Many problems such as  pedestrian detection~\cite{app6_pedestrian}, traffic surveillance~\cite{app5_traffic} require the segmentation of specific objects e.g. persons or cars. Other domains include satellite imagery~\cite{app7_satellite,app8_satellite}, guidance systems in defense~\cite{app9_military}, forensics such as face~\cite{app10_face}, iris~\cite{app11_iris} and fingerprint~\cite{app12_fingerprint} recognition. Generally traditional methods such as histogram thresholding~\cite{met1_hist}, hybridization~\cite{met2_hist,met3_hybr} feature space clustering~\cite{met4_feat}, region-based approaches~\cite{survey9_region}, edge detection approaches~\cite{met5_edge}, fuzzy approaches~\cite{met6_fuzzy}, entropy-based approaches~\cite{met7_entropy}, neural networks (Hopfield neural network~\cite{met10_hopfield}, self-organizing maps~\cite{met12_sofm}), physics-based approaches~\cite{met13_physics} etc. are used popularly in this purpose. However, such feature-based approaches have a common bottleneck that they are dependent on the quality of feature extracted by the domain experts. Generally, humans are bound to miss latent or abstract features  for image segmentation. On the other hand, deep learning in general addresses this issue of automated feature learning. In this regard one of the most common technique in computer vision was introduced soon by the name of convolutional neural networks~\cite{deep_lenet} that learned a cascaded set of convolutional kernels through backpropagation ~\cite{hist_backprop_rumelhart1986learning}. Since then, it has been improved significantly with features like layer-wise training~\cite{deep_layerwise_rbm}, rectified linear activations~\cite{deep_relu}, batch normalization~\cite{deep_batchnorm}, auxiliary classifiers~\cite{deep_auxiliary}, atrous convolutions~\cite{seg_atrous}, skip connections~\cite{deep_resnet}, better optimization techniques~\cite{deep_adam} and so on. With all these there was a large number of new types of image segmentation techniques as well. Various such techniques drew inspiration from popular networks such as AlexNet~\cite{deep_alexnet}, convolutional autoencoders~\cite{deep_stackedAE}, recurrent neural networks~\cite{deep_rnn}, residual networks~\cite{deep_resnet} and so on.

\section{Motivation}
There have been many reviews and surveys regarding the traditional technologies associated with image segmentation~\cite{survey1,survey3}. While some of them specialized in application areas~\cite{survey4_color,survey5_medical,survey8_medical}, while other focused on specific types of algorithms~\cite{survey10_salient,survey7_salient,survey9_region}. With arrival of deep learning techniques many new classes of image segmentation algorithms have surfaced. Earlier studies~\cite{survey_deep_old} have shown the potential of deep learning based approaches. There have been more recent studies ~\cite{survey_deep_small} which cover a number of methods and compare them on the basis of their reported performance.	The work of Garcia et al.~\cite{survey6_deep} lists a variety of deep learning based segmentation techniques. They have tabulated the performance of various state of the art networks on several modern challenges. The resources are incredibly useful for understanding the current state-of-the-art in this domain. While knowing the available methods is quite useful to develop products, however, to contribute to this domain as a researcher, one needs to understand the underlying mechanics of the methods that make them confident. In the present work, our main motivation is to answer the question why the methods are designed in a way they are. Understanding the mechanics of modern techniques would make it easier to tackle new challenges and develop better algorithms. Our approach carefully analyses each method to understand why they succeed at what they do and also why they fail for certain problems. Being aware of pros and cons of such method new designs can be initiated that reaps the benefits of the pros and overcomes the cons. We recommend the works of Alberto Garcia-Garcia~\cite{survey6_deep}  to get an overview of some of the best image segmentation techniques using deep learning while our focus would be to understand why, when and how these techniques perform on various challenges. 
\begin{figure*}[htbp]
	\centering
	\includegraphics[width=0.8\textwidth]{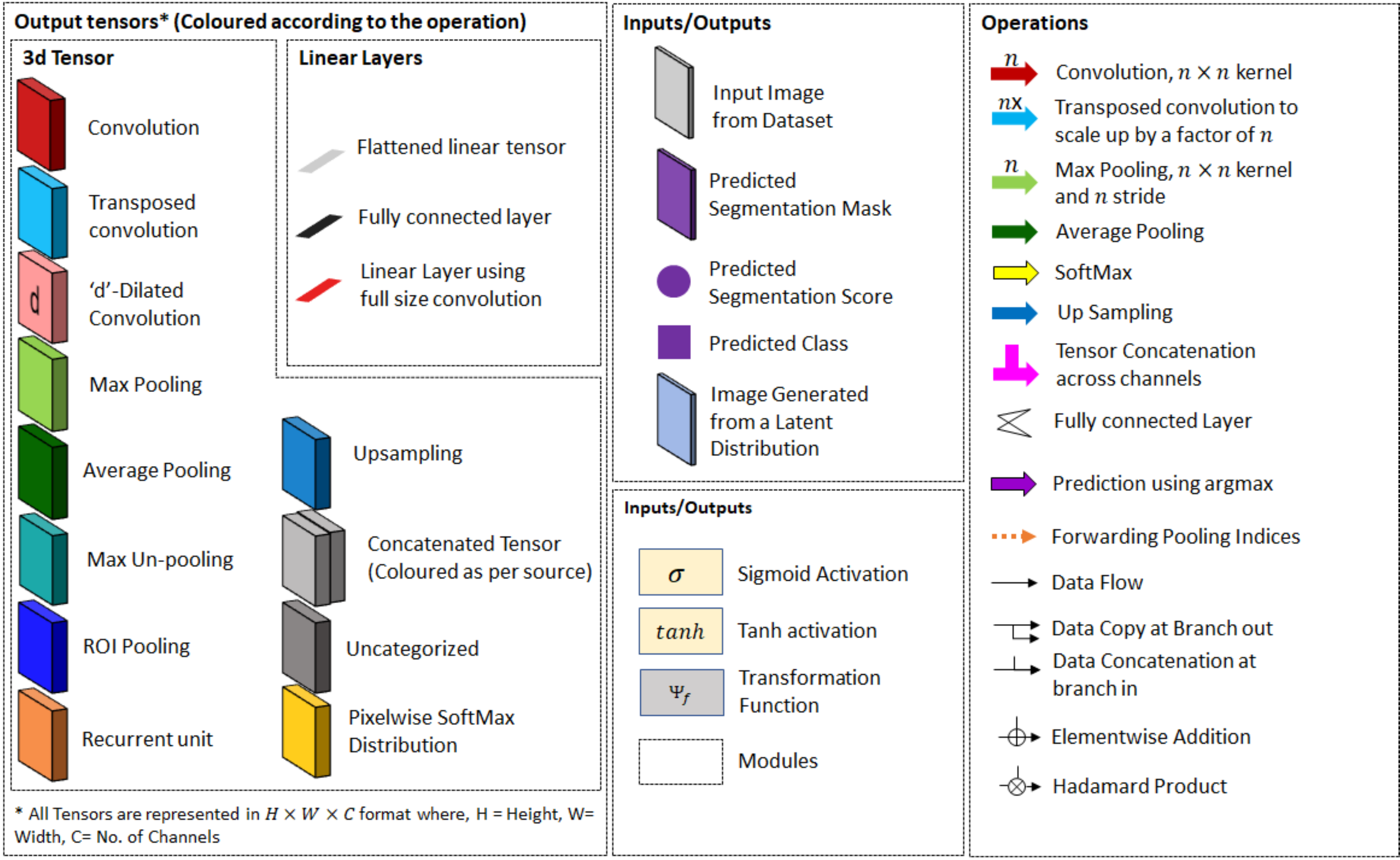}
	\caption{Legends for subsequent diagrams of popular deep learning architectures}
	\label{fig:legends}
\end{figure*}
\subsection{Contribution}
The paper has been designed in a way such that new researchers reap the most benefits. Initially some of the traditional techniques have been discussed to uphold the frameworks before the deep learning era. Gradually the various factors governing the onset of deep learning has been discussed so that readers have a good idea of the current direction in which machine learning is progressing. In the subsequent sections the major deep learning algorithms have been briefly described in a generic way to establish a clearer concept of the procedures in the mind of the readers. The image segmentation algorithms discussed thereafter have been categorized into the major families of algorithms that governed the last few years in this domain. The concepts behind all the major approaches have been explained through a very simple language with minimum amount of complicated mathematics. Almost all the diagrams corresponding to major networks have been drawn using a common representational format as shown in fig.~\ref{fig:legends}. The various approaches that have been discussed comes with different representations for architectures. The unified representation scheme allows the user to understand the fundamental similarities and differences between networks. Finally, the major application areas have been discussed to help new researchers pursue a field of their choice.
\section{Impact of Deep Learning on Image Segmentation}
The development of deep learning algorithms like convolutional neural networks or deep autoencoders not only affected typical tasks like object classification but are also efficient in other related tasks like object detection, localization, tracking, or as in this case image segmentation.
\subsection{Effectiveness of convolutions for segmentation}
As an operation convolution can be simply defined as the function that performs a sum-of-product between kernel weights and input values while convoluting the smaller kernel over a larger image. For a typical image with $k$ channels we can convolute a smaller sized kernel with $k$ channels along the $x$ and $y$ direction to obtain an output in the format of a 2 dimensional matrix. It has been observed that after training a typical CNN the convolutional kernels tend to generate activation maps with respect to certain features of the objects~\cite{deep_visualizing}. Given the nature of activations, it can be seen as segmentation masks of object specific features. Hence the key to generating requirement specific segmentation is already embedded within this output activation matrices. Most of the image segmentation algorithm uses this property of CNNs to somehow generate the segmentation masks as required to solve the problem. As shown below in fig. ~\ref{fig:maps}, the earlier layers capture local features like the contour or a small part of an object. In the later layers more global features are activated such as field, people or sky. It can also be noted from this figure that the earlier layers show sharper activations as compared to the later ones.
\begin{figure}[htbp]
	\centering
	\includegraphics[width=0.75\textwidth]{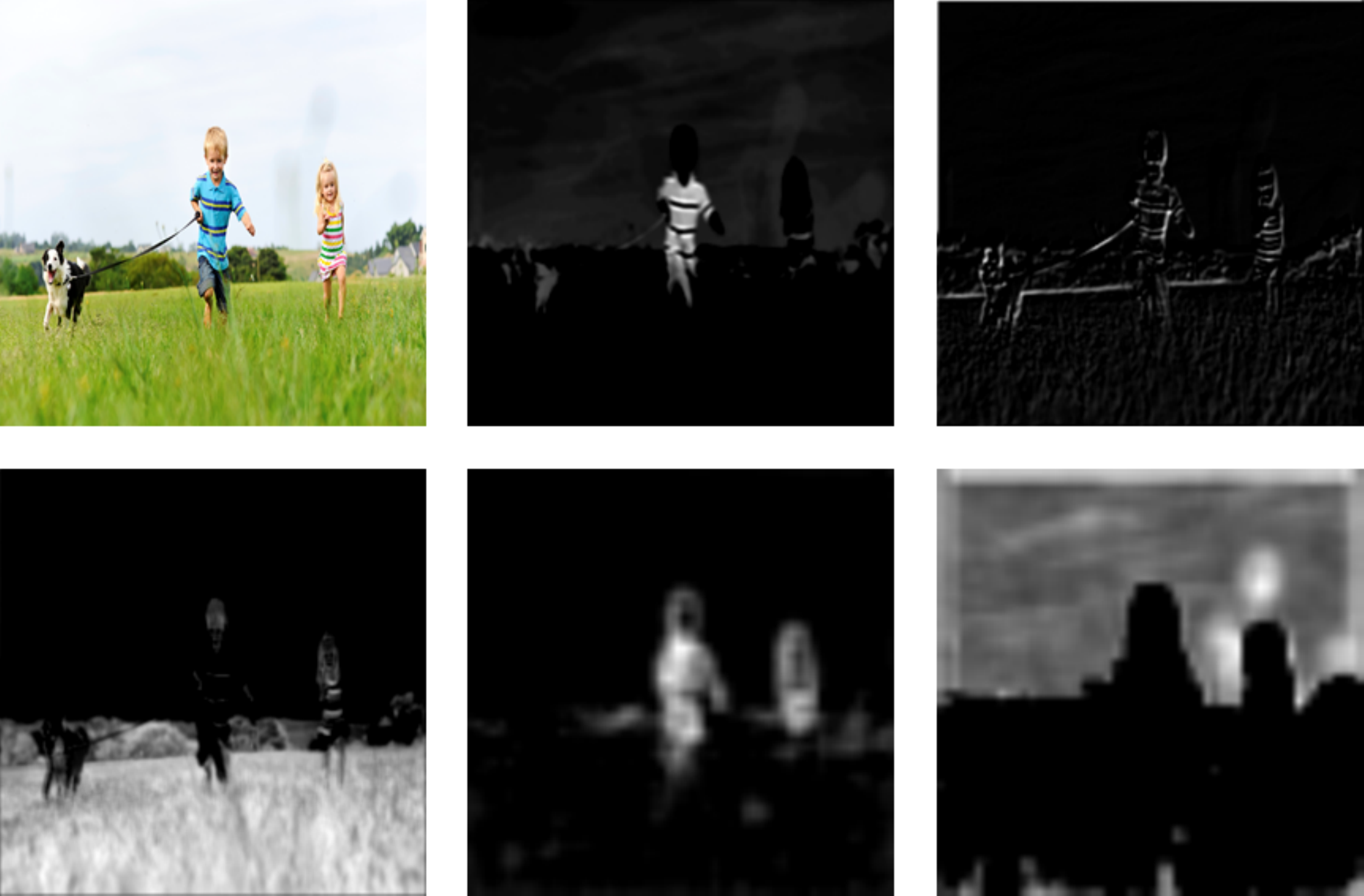}
	\caption{Input image and sample activation maps from a typical CNN. (Top row) Input image and two activation maps from earlier layers showing part objects like t-shirts and features like contours. (Bottom row) shows activation maps from later layers with more meaningful activations like fields, people and sky respectively}
	\label{fig:maps}
\end{figure}
\subsection{Impact of larger and more complex datasets}
The second impact that deep learning brought to the world of image segmentation is the plethora of datasets, challenges and competitions. These factors encouraged researchers across the world to come up with various state-of-the-art technologies to implement segmentation across various domains. A list of many such datasets have been provided in table \ref{tab:dataset}

\begin{table}
	\centering
	\caption{A list of various datasets in the image segmentation domain}
	\label{tab:dataset}
	\resizebox{\textwidth}{!}{%
	\begin{tabular}{l l l}
		\hline
		\textbf{Category} 		& \textbf{Dataset} \\		
		\hline
		\hline													
		Natural				& Berkeley Segmentation Dataset ~\cite{data_BSDS500} \\
		Scenes				& PASCAL VOC ~\cite{data_pascal} \\
		& Stanford Background Dataset ~\cite{data_StanfordBD} \\
		& Microsoft COCO ~\cite{data_COCO} \\
		& MIT Scene parsing data(ADE20K) ~\cite{data_ade20k,data_ade20k_2} \\
		& Semantic Boundaries Dataset ~\cite{data_semanticBD} \\
		& Microsoft Research Cambridge Object Recognition Image Database (MSRC) ~\cite{data_msrc21} \\
		\hline
		Video				& Densely Annotated Video Segmentation(DAVIS) ~\cite{data_DAVIS} \\
		Segmentation		& Video Segmentation Benchmark(VSB100) ~\cite{data_vsb100}\\
		Dataset				& YouTube-Video object Segmentation ~\cite{data_youtubevos}\\
		\hline	
		Autonomous			& Cambridge-driving Labeled Video Database (CamVid) ~\cite{data_camvid} \\
		Driving				& Cityscapes: Semantic Urban Scene Understanding ~\cite{data_cityscapes} \\
		& Mapillary Vistas Dataset ~\cite{data_mapillary} \\
		& SYNTHIA: Synthetic collection of Imagery and Annotations ~\cite{data_synthia} \\
		& KITTI Vision Benchmark Suite ~\cite{data_kitti} \\
		& Berkeley Deep Drive ~\cite{data_BDD} \\
		& India Driving Dataset(IDD) ~\cite{data_idd}\\
		\hline
		Aerial				& Inria Aerial Image Labeling Dataset ~\cite{data_inria} \\
		Imaging				& Aerial Image Segmentation Dataset~\cite{data_aerial} \\
		& ISPRS Dataset collection ~\cite{data_isprs} \\
		& Google Open Street Map ~\cite{data_googleosm} \\
		& DeepGlobe ~\cite{data_deepglobe} \\
		\hline
		Medical				& DRIVE:Digital Retinal Images for Vessel Extraction ~\cite{data_drive} \\
		Imaging				& Sunnybrook Cardiac Data ~\cite{data_scd} \\
		& Multiple Sclerosis Database ~\cite{data_multiplesclerosis,data_sclerosis} \\
		& IMT: Intima Media Thickness Segmentation Dataset ~\cite{data_imt} \\
		& SCR: Segmentation in Chest Radiographs ~\cite{data_SCR} \\
		& BRATS: Brain Tumor Segmentation ~\cite{data_BRATS} \\
		& LITS: Liver Tumour Segmentation ~\cite{data_lits} \\
		& BACH: Breast Cancer Histology ~\cite{data_bach} \\
		& IDRiD: Indian Diabetic Retinopathy Image Dataset ~\cite{data_idrid} \\
		& ISLES: Ischemic Stroke Lesion Segmentation ~\cite{data_isles} \\
		\hline
		Saliency			& MSRA Salient Object Database ~\cite{data_msra10k} \\
		Detection			& ECSSD: Extended Complex Scene Saliency Dataset ~\cite{data_eccsd} \\
		& PASCAL-S DATASET ~\cite{data_salobj} \\
		& THUR15K: Group Saliency in Image ~\cite{data_thur15k} \\
		& JuddDB: MIT saliency benchmark ~\cite{data_judddb} \\
		& DUT-OMRON Image Dataset ~\cite{data_dutomron} \\
		\hline
		
		Scene Text 		    & KAIST Scene Text Database ~\cite{data_kaist} \\
		Segmentation		& COCO-Text ~\cite{data_cocotext} \\
		& SVT: Street View Text Dataset ~\cite{data_SVT} \\
		\hline		
	\end{tabular} }%
\end{table}
\section{Image Segmentation using Deep Learning}
As explained before, convolutions are quite effective in generating semantic activation maps that has components which inherently constitute various semantic segments. Various methods have been implemented to make use of these internal activations to segment the images. A summary of major deep learning based segmentation algorithms are provided in table \ref{tab:methods} along with brief description of their major contribution.
\begin{table}[]
	\centering
	\caption{A summary of major deep learning based segmentation algorithms. Abbreviations: S: Supervised, W: Weakly supervised, U: Unsupervised, I: Interactive, P: Partially Supervised, SO: Single objective optimization, MO: Multi objective optimization, AD: Adversarial Learning, SM: Semantic Segmentation, CL: Class specific Segmentation, IN: Instance Segmentation, RNN: Recurrent Modules, E-D: Encoder Decoder Architecture}
	\label{tab:methods}
	\renewcommand\arraystretch{1.2}
	\resizebox{\textwidth}{!}{%
	\begin{tabular}{|l|c|ccccc|ccc|ccc|cc|l|}
		\hline
		\multicolumn{1}{|c|}{\multirow{2}{*}{\textbf{Method}}} & \multicolumn{1}{c|}{\multirow{2}{*}{\textbf{Year}}} & \multicolumn{5}{c|}{\textbf{Supervision}} & \multicolumn{3}{c|}{\textbf{Learning}} & \multicolumn{3}{c|}{\textbf{Type}} & \multicolumn{2}{c|}{\textbf{Modules}} & \multicolumn{1}{c|}{\multirow{2}{*}{\textbf{Description}}} \\ \cline{3-15}
		\multicolumn{1}{|c|}{} & \multicolumn{1}{c|}{} & S & W & U & I & \multicolumn{1}{c|}{P} & SO & MO & \multicolumn{1}{c|}{AD} & SM & CL & \multicolumn{1}{c|}{IN} & RNN & \multicolumn{1}{c|}{E-D} & \multicolumn{1}{c|}{} \\ \hline
		Global Average Pooling & 2013 &  & \checkmark &  &  &  & \checkmark &  &  &  & \checkmark &  &  &  & Object specific soft segmentation \\
		DenseCRF & 2014 &  &  &  &  & \checkmark & \checkmark &  &  & \checkmark &  &  &  &  & Using CRF to boost segmentation \\
		FCN & 2015 & \checkmark &  &  &  &  & \checkmark &  &  & \checkmark &  &  &  &  & Fully convolutional layers \\
		DeepMask & 2015 & \checkmark &  &  &  &  &  & \checkmark &  &  & \checkmark &  &  &  & Simultaneous learning for segmentation and classification \\
		U-Net & 2015 & \checkmark &  &  &  &  & \checkmark &  &  & \checkmark &  &  &  & \checkmark & Encoder-Decoder with multiscale feature concatenation \\
		SegNet & 2015 & \checkmark &  &  &  &  & \checkmark &  &  & \checkmark &  &  &  & \checkmark & Encoder-Decoder with forwarding pooling indices \\
		CRF as RNN & 2015 & \checkmark &  &  &  &  &  & \checkmark &  & \checkmark &  &  & \checkmark &  & Simulating CRFs as trainable RNN modules \\
		Deep Parsing Network & 2015 & \checkmark &  &  &  &  &  & \checkmark &  &  &  &  &  &  & Using unshared kernels to incorporate higher order dependency \\
		BoxSup & 2015 &  & \checkmark &  &  &  &  &  &  & \checkmark &  &  &  &  & Using bounding box for weak supervision \\
		SharpMask & 2016 & \checkmark &  &  &  &  &  & \checkmark &  &  & \checkmark &  &  & \checkmark & Refined Deepmask with multi layer feature fusion \\
		Attention to Scale & 2016 & \checkmark &  &  &  &  & \checkmark &  &  & \checkmark &  &  &  &  & Fusing features from multi scale inputs \\
		Semantic Segmentation & 2016 & \checkmark &  &  &  &  &  &  & \checkmark & \checkmark &  &  &  &  & Adversarial training for image segmentation \\
		Conv LSTM and Spatial Inhibition & 2016 & \checkmark &  &  &  &  &  & \checkmark &  &  &  & \checkmark & \checkmark &  & Using spatial inhibition for instance segmentation \\
		JULE & 2016 &  &  & \checkmark &  &  &  & \checkmark &  & \checkmark &  &  & \checkmark &  & Joint unsupervised learning for segmentation \\
		ENet & 2016 & \checkmark &  &  &  &  & \checkmark &  &  & \checkmark &  &  &  &  & Compact network for realtime segmentation \\
		Instance aware segmentation & 2016 & \checkmark &  &  &  &  &  & \checkmark &  &  &  & \checkmark &  &  & Multi task approach for instance segmentation \\
		Mask RCNN & 2017 & \checkmark &  &  &  &  &  & \checkmark &  & \checkmark &  &  &  &  & Using region propsal network for segmentation \\
		Large Kernel Matters & 2017 & \checkmark &  &  &  &  & \checkmark &  &  & \checkmark &  &  &  & \checkmark & Using larger kernels for learning complex features \\
		RefineNet & 2017 & \checkmark &  &  &  &  & \checkmark &  &  & \checkmark &  &  &  & \checkmark & Multi path refinement module for fine segmentation \\
		PSPNet & 2017 & \checkmark &  &  &  &  & \checkmark &  &  & \checkmark &  &  &  &  & Multi scale pooling for scale agnostic segmentation \\
		Tiramisu & 2017 & \checkmark &  &  &  &  & \checkmark &  &  & \checkmark &  &  &  & \checkmark & DenseNet 121 feature extractor \\
		Image to Image Translation & 2017 & \checkmark &  &  &  &  &  &  & \checkmark & \checkmark &  &  &  & \checkmark & Conditional GAN for translation image to segment maps \\
		Instance Segmentation with attention & 2017 & \checkmark &  &  &  &  &  & \checkmark &  &  &  & \checkmark & \checkmark &  & Attention modules for image segmentatoin \\
		W-Net & 2017 &  &  & \checkmark &  &  &  & \checkmark &  & \checkmark &  &  &  & \checkmark & Unsupervised segmentation using normalized cut loss \\
		PolygonRNN & 2017 &  &  &  & \checkmark &  & \checkmark &  &  & \checkmark &  &  & \checkmark &  & Generating contours by RNN \\
		Deep Layer Cascade & 2017 & \checkmark &  &  &  &  & \checkmark &  &  & \checkmark &  &  &  &  & Multi level approach to handle pixels of different complexity \\
		Spatial Propagation Network & 2017 & \checkmark &  &  &  &  & \checkmark &  &  & \checkmark &  &  &  &  & Refinement using linear label propagation \\
		DeepLab & 2018 & \checkmark &  &  &  &  & \checkmark &  &  & \checkmark &  &  &  &  & Atrous convolution, Spatial pooling pyramid, DenseCRF \\
		SegCaps & 2018 & \checkmark &  &  &  &  &  & \checkmark &  &  &  &  &  &  & Capsule Networks for Segmentation \\
		Adversarial Collaboration & 2018 &  &  & \checkmark &  &  &  & \checkmark &  &  &  &  &  &  & Adversarial collaboration between multiple networks \\
		Superpixel Supervision & 2018 &  &  & \checkmark &  &  &  & \checkmark &  &  &  &  &  &  & Using superpixel refinement as supervisory signals \\
		Deep Extreme Cut & 2018 &  &  &  & \checkmark &  & \checkmark &  &  & \checkmark &  &  &  &  & Using extreme points for interactive segmentation \\
		Two Stream Fusion & 2019 &  &  &  & \checkmark &  & \checkmark &  &  & \checkmark &  &  &  &  & Using image stream and interaction stream simultaneously \\
		SegFast & 2019 & \checkmark &  &  &  &  & \checkmark &  &  & \checkmark &  &  &  & \checkmark & Using depth-wise separable convolution in SqueezeNet encoder \\ \hline
	\end{tabular}}%
\end{table}

\subsection{Convolutional Neural Networks}
Convolutional neural networks being one of the most commonly used methods in computer vision has adopted many simple modifications to perform well in segmentation tasks as well. 
\begin{figure}[htbp]
	\centering
	\includegraphics[width=0.6\textwidth]{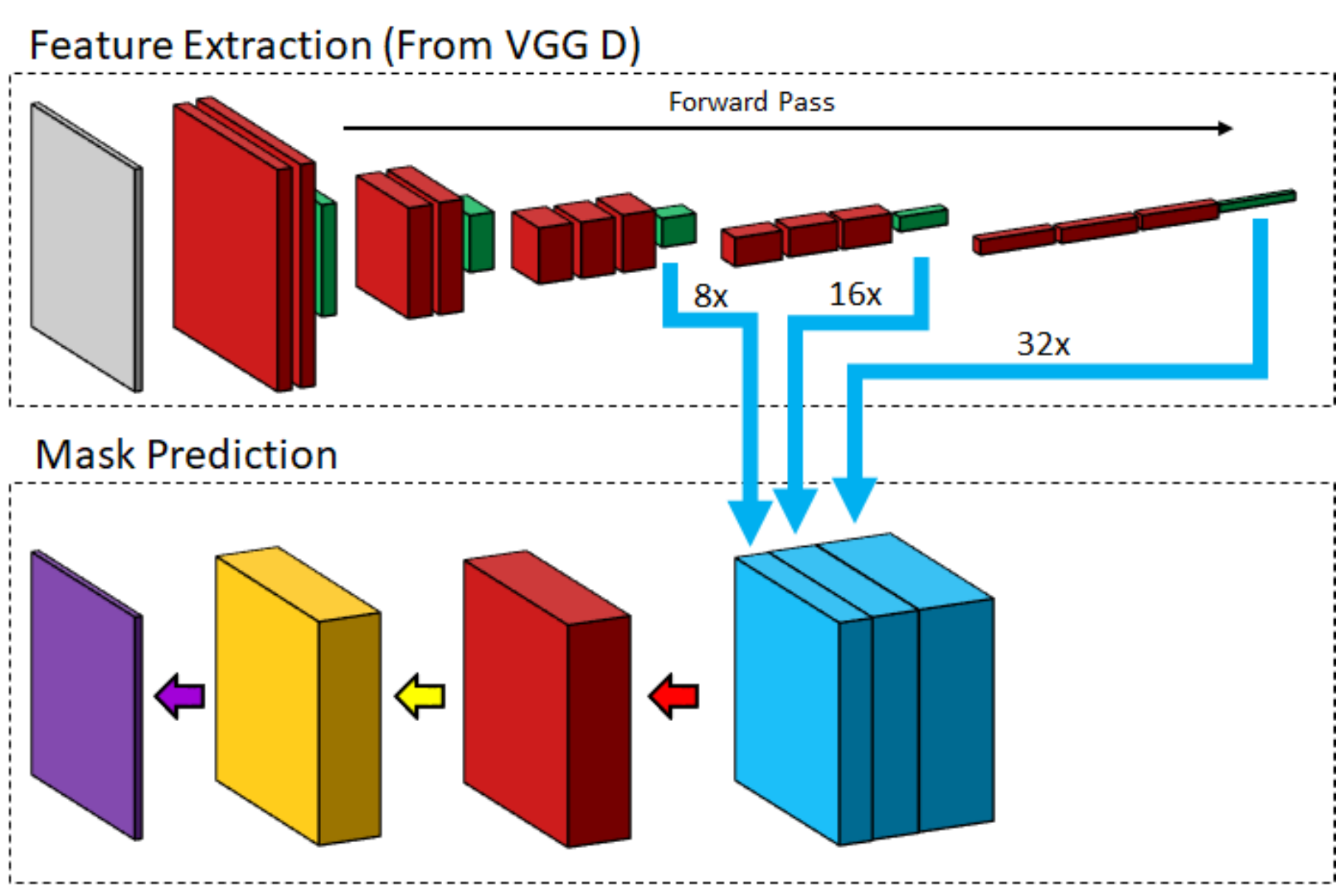}
	\caption{A Fully convolutional network with image segmentation with concatenated multi-scale features}
	\label{fig:fcn}
\end{figure}
\subsubsection{Fully convolutional layers}
Classification tasks generally require a linear output in the form of a probability distribution over the number of classes. To convert volumes of 2 dimensional activation maps into linear layers they were often flattened. The flattened shape allowed the execution of fully connected networks to obtain the probability distribution. However, this kind of reshaping loses the spatial relations among the pixels in the image. In a fully convolutional neural network(FCN) ~\cite{seg_fcn} the output of the last convolutional block is directly used for a pixel level classification. FCNs were first implemented on the PASCAL VOC 2011 segmentation dataset\cite{data_pascal} and achieved a pixel accuracy of 90.3\% and a mean IOU of 62.7\%. Another way to avoid fully connected linear layers is the use of a full size average pooling to convert a set of 2 dimensional activation maps to a set of scalar values. As these pooled scalars are connected to the output layers, the weights corresponding to each class may be used to perform weighted summation of the corresponding activation maps in the previous layers. This process called Global Average Pooling(GAP) ~\cite{seg_gap} can be directly used on various trained networks like residual network to find object specific activation zones which can be used for pixel level segmentation. The major issues with algorithm such as this is the loss of sharpness due to the intermediate sub-sampling operations. Sub-sampling is a common operation in convolutional neural networks to increase the sensory area of kernels. What it means is that as the activations maps reduces in size in the subsequent layers, the kernels convoluting over them actually corresponds to a larger area in the original image. However, it reduces the image size in the process, which when up-sampled to original size loses sharpness. Many approaches have been implemented to handle this issue. For fully convolutional models, skip connections from preceding layers can be used to obtain sharper versions of the activations from which finer segments can be chalked out (Refer fig. \ref{fig:fcn}). Another work showed how the usage of high dimensional kernels to capture global information with FCN models created better segmentation masks~\cite{seg_large}. Segmentation algorithms can also be treated as boundary detection technique. convolutional features are also very useful from that perspective~\cite{convorientedboundary}. While earlier layers can provide fine details, later layers focus more on the coarser boundaries.

\begin{figure}[htbp]
	\centering
	\includegraphics[width=0.9\textwidth]{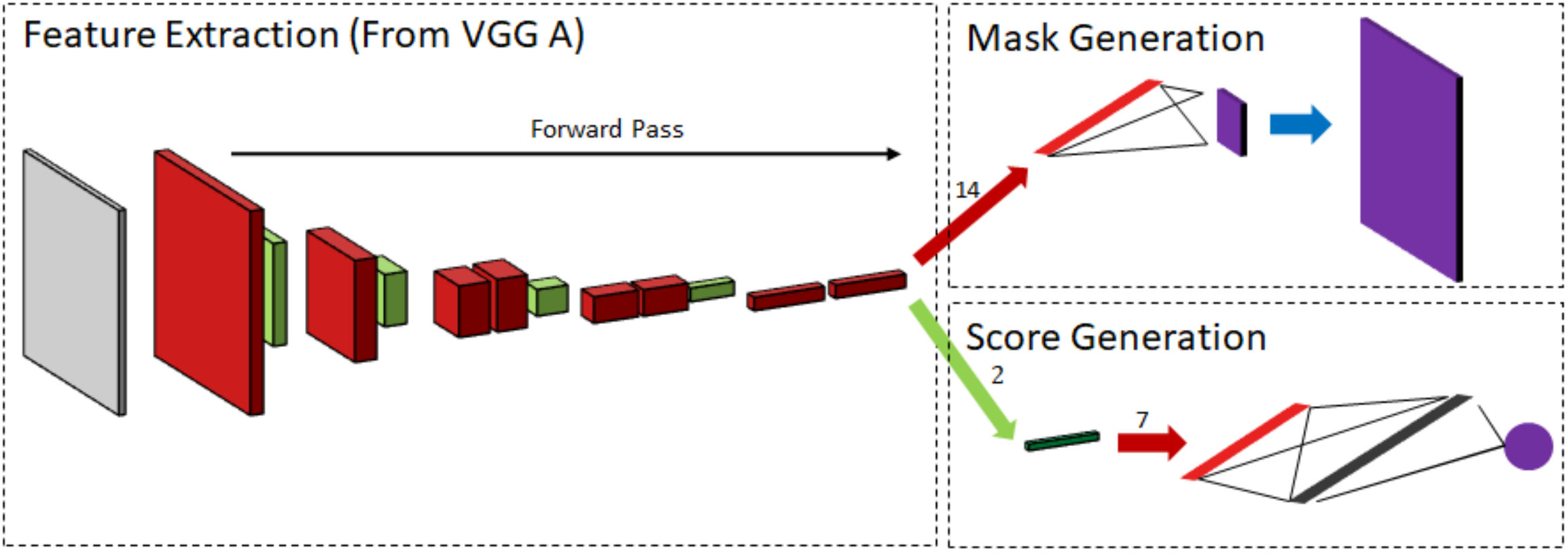}
	\caption{The Deepmask Network}
	\label{fig:deepmask}
\end{figure}

\paragraph*{DeepMask and SharpMask} DeepMask \cite{seg_deepmask} was a name given to a project at Facebook AI Research (FAIR) related to image segmentation. It exhibited the same school of thought as FCN models except that the model was capable of multi-tasking (Refer fig. \ref{fig:deepmask}).	
\begin{figure}[htbp]
	\centering
	\includegraphics[width=0.8\textwidth]{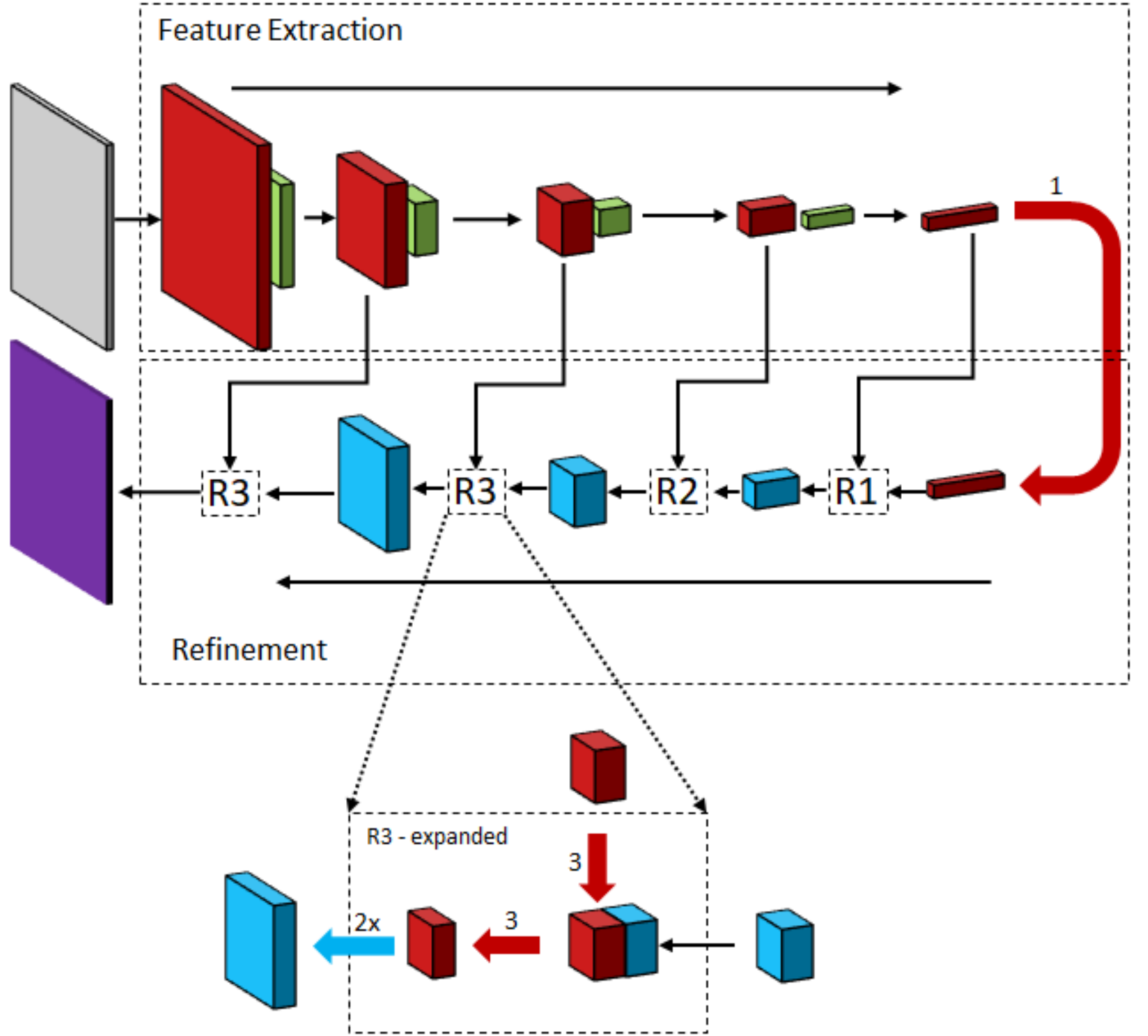}
	\caption{The Sharpmask Network}
	\label{fig:sharpmask}
\end{figure} It had two main branches coming out of a shared feature representation. One of them created a pixel level classification of or a probabilistic mask for the central object and the second branch generated a score corresponding to the object recognition accuracy. The network coupled with sliding windows of sixteen strides to create segments of objects at various locations of the image, whereas the score helped in identifying which of the segments were good. The network was further upgraded in SharpMask~\cite{seg_sharpmask}, where probabilistic masks from each layer were combined in top-down fashion using convolutional refinements at every steps to generate high resolution masks (Refer fig. \ref{fig:sharpmask}). The sharpmask scored an average recall of 39.3 which beats deepmask, which scored 36.6 on the MS COCO Segmentation Dataset.

\subsubsection{Region proposal networks}
Another similar wing that started developing with image segmentation was object localization. Task such as this involved locating specific objects in images. Expected outputs for such problems is normally a set of bounding boxes corresponding to the queried objects. Though strictly stating, some of these algorithms do not address image segmentation problems, however their approaches are of relevance to this domain.

\begin{figure}[htbp]
	\centering
	\includegraphics[width=0.75\textwidth]{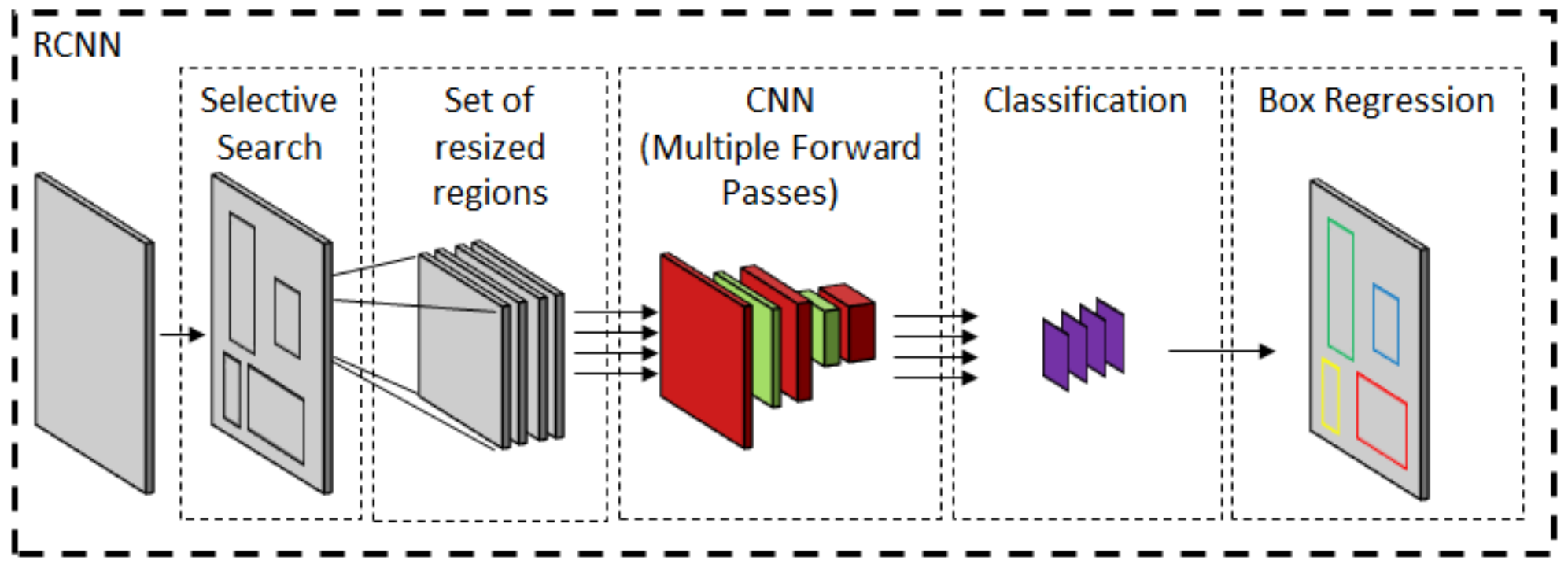}
	\includegraphics[width=0.75\textwidth]{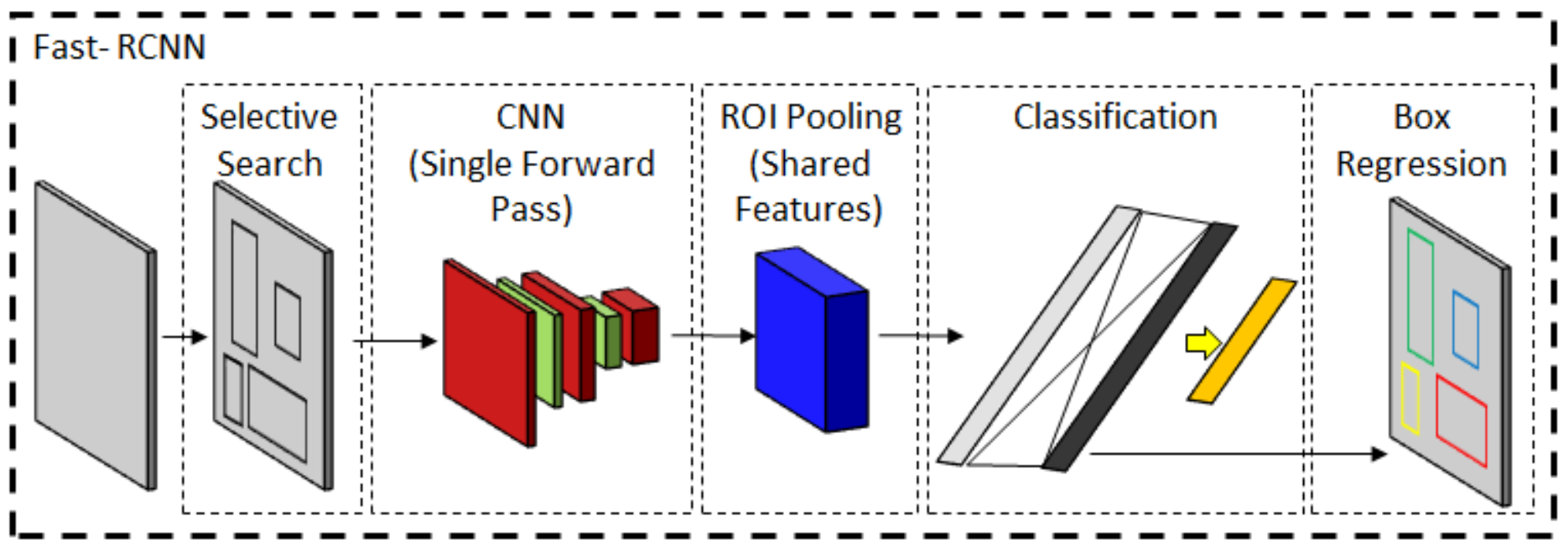}
	\includegraphics[width=0.75\textwidth]{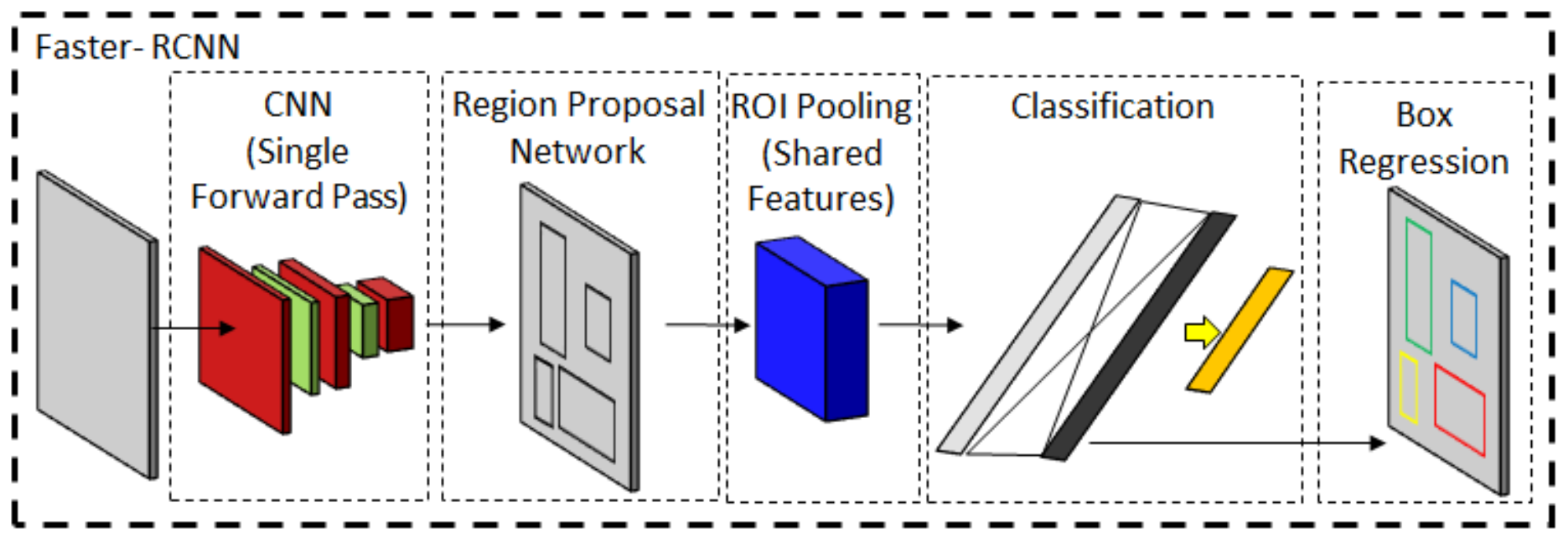}
	\includegraphics[width=0.75\textwidth]{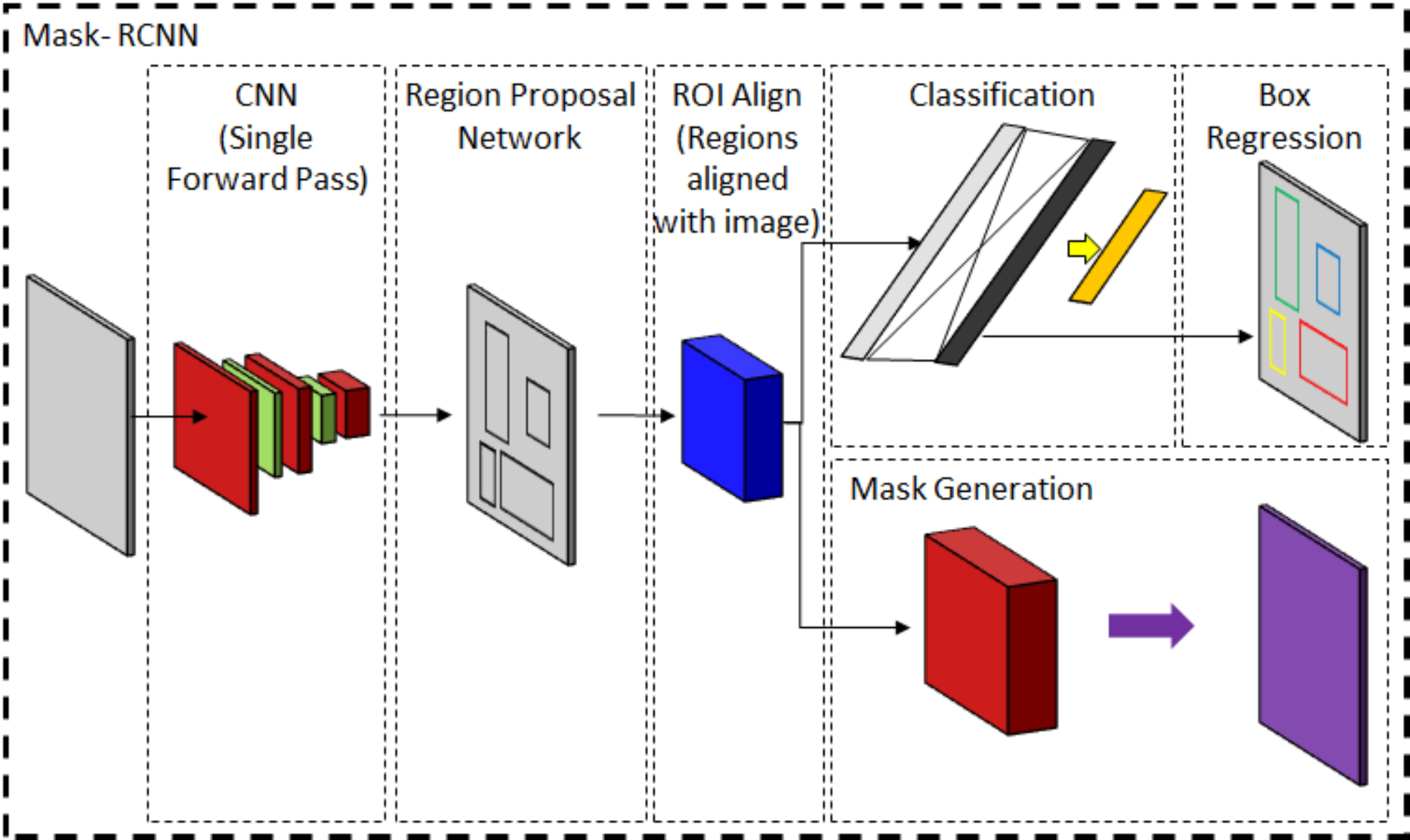}		
	\caption{The RCNN Family of localization and segmentation networks}
	\label{fig:rcnn}
\end{figure}

\paragraph*{RCNN (Region-based Convolutional Neural Networks)}
The introduction of the CNNs raised many new questions in the domain of computer vision. One of them primarily being whether a network like AlexNet can be extended to detect the presence of more than one object. Region-based-CNN~\cite{seg_rcnn} or more commonly known as R-CNN used selective search technique to propose probable object regions and performed classification on the cropped window to verify sensible localization based on the output probability distribution. Selective search technique~\cite{met_selective,met_selectivesearch_Segmentation} analyses various aspects like texture, color, or intensities to cluster the pixels into objects. The bounding boxes corresponding to these segments are passed through classifying networks to short-list some of the most sensible boxes. Finally, with a simple linear regression network tighter co-ordinate can be obtained. The main downside of the technique is its computational cost. The network needs to compute a forward pass for every bounding box proposition. The problem with sharing computation across all boxes was that the boxes were of different sizes and hence uniform sized features were not achievable. In the upgraded Fast R-CNN~\cite{seg_fastrcnn}, ROI (Region of Interest) Pooling was proposed in which region of interests were dynamically pooled to obtain a fixed size feature output. Henceforth, the network was mainly bottlenecked by the selective search technique for candidate region proposal. In Faster-RCNN ~\cite{seg_fasterrcnn}, instead of depending on external features, the intermediate activation maps were used to propose bounding boxes, thus speeding up the feature extraction process. Bounding boxes are representative of the location of the object, however they do not provide pixel-level segments. The Faster R-CNN network was extended as Mask R-CNN~\cite{seg_maskrcnn} with a parallel branch that performed pixel level object specific binary classification to provide accurate segments. With Mask-RCNN an average precision of 35.7 was attained in the COCO\cite{data_COCO} test images. The family of RCNN algorithms have been depicted in fig.\ref{fig:rcnn}. Region proposal networks have often been combined with other networks ~\cite{seg_fcnrcnn, seg_rfcn} to give instance level segmentations. RCNN was further improved under the name of HyperNet ~\cite{seg_hypernet}  by using features from multiple layers of the feature extractor. Region proposal networks have also been implemented for instance specific segmentation as well. As mentioned before object detection capabilities of approaches like RCNN are often coupled with segmentation models to generate different masks for different instances of the same object\cite{dai_instance}.
\subsubsection{DeepLab}
\label{sec:deeplab}
While pixel level segmentation was effective, two complementing issues were still affecting the performance. Firstly, smaller kernel sizes failed to capture contextual information. In classification problems, this is handled using pooling layers that increases the sensory area of the kernels with respect to the original image. But in segmentation that reduces the sharpness of the segmented output. Alternative usage of larger kernels tend to be slower due to significanty larger number of trainable parameters. To handle this issue the DeepLab~\cite{seg_deeplab,seg_deeplabv3} family of algorithms demonstrated the usage of various methodologies like atrous convolutions~\cite{seg_atrous}, spatial pooling pyramids~\cite{seg_spatial_pyramid_pooling} and fully connected conditional random fields~\cite{seg_densecrf} to perform image segmentation with great efficiency. The DeepLab algorithm was able to attain a meanIOU of 79.7 on the PASCAL VOC 2012 dataset\cite{data_pascal}.
\begin{figure}[htbp]
	\centering
	\includegraphics[width=0.45\textwidth]{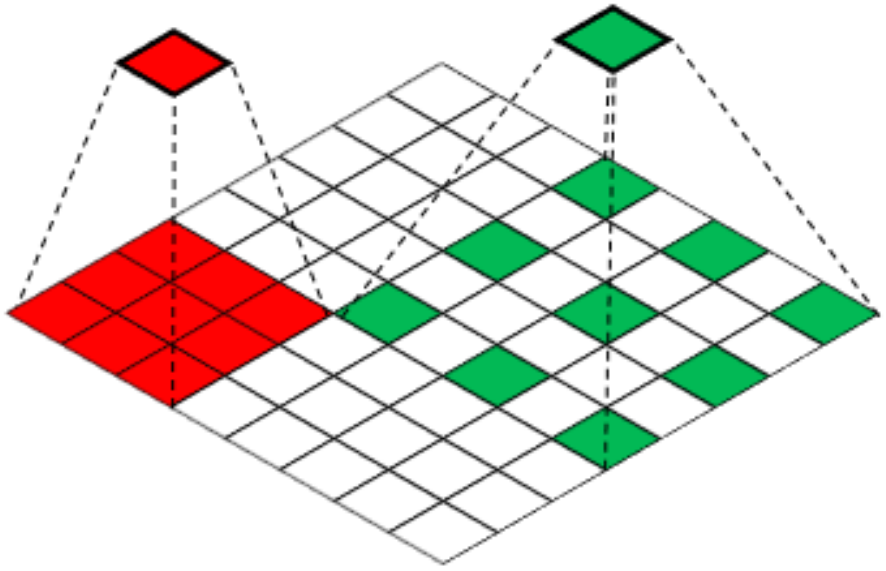}
	\caption{Normal convolution(red) vs. Atrous or Dilated convolution(green)}
	\label{fig:atrous}
\end{figure}
\paragraph*{Atrous/Dilated Convolution}
The size of the convolution kernels in any layer determine the sensory response area of the network. While smaller kernels extract local information, larger kernels try to focus on more contextual information. However, larger kernels normally comes with more number of parameters. For example to have a sensory region of $6\times 6$, one must have $36$ neurons. To reduce the number of parameters in the CNN, the sensory area is increased in higher layers through techniques like pooling. Pooling layers reduce the  size of the image. When an image is pooled by a $2 \times 2$ kernel with a stride of two, the size of the image reduces by 25\%. A kernel with an area of $3\times 3$ corresponds to a larger sensory area of $6\times 6 $ in the original image. However, unlike before now only $18$ neurons ($9$ for each layer) are needed in the convolution kernel. In case of segmentation, pooling creates new problems. The reduction in the image size results in loss of sharpness in generated segments as the reduced maps are scaled up to image size. To deal with these two issues simultaneously, dilated or atrous convolutions play a key role. Atrous/Dilated convolutions increase the field of view without increasing the number of parameters. As shown in fig.\ref{fig:atrous} a $3\times 3$ kernel with a dilation factor of $1$ can act upon an area of $5 \times 5$ in the image. Each row and column of the kernel has three neurons which is multiplied with intensity values in the image which separated by the dilation factor of $1$. In this way the kernels can span over larger areas while keeping the number of neurons low and also preserving the sharpness of the image. Besides the DeepLab algorithms, atrous convolutions ~\cite{seg_atrous_encdec} have also been used with auto encoder based architectures.
\begin{figure}[htbp]
	\centering
	\includegraphics[width=0.3\textwidth]{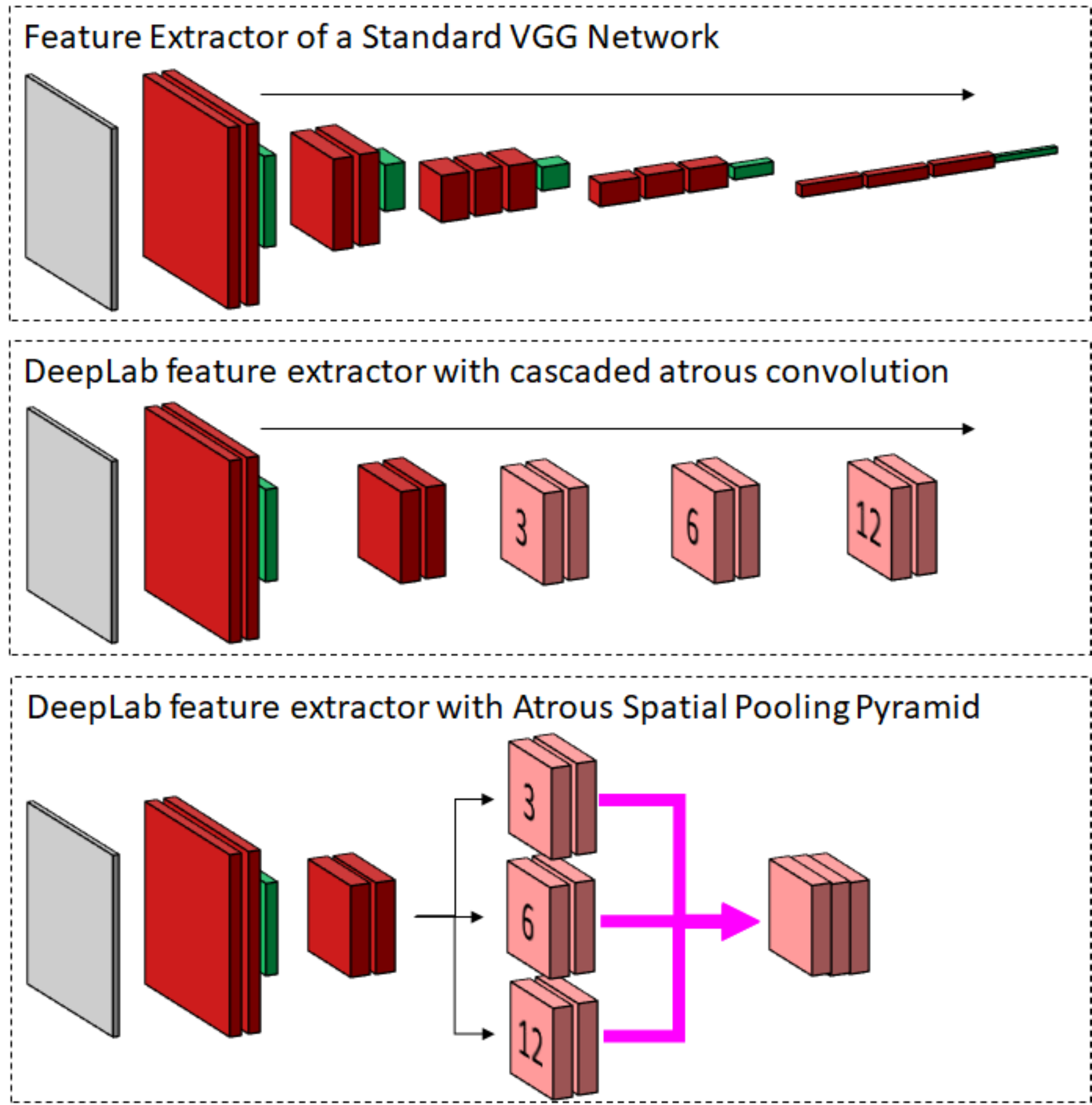}
	\caption{DeepLab Architecture as compared to a standard VGG net(top) along with cascaded atrous convolutions (middle) and atrous spatial pooling pyramid(bottom) }
	\label{fig:deeplab}
\end{figure}	
\paragraph*{Spatial Pyramid Pooling}
Spatial pyramid pooling~\cite{seg_spatial_pyramid_pooling} was introduced in R-CNN where ROI pooling showed the benefit of using multi-scale regions for object localization. However, in DeepLab, atrous convolutions were preferred over pooling layers for changing field of view or sensory area. To imitate the effect of ROI pooling, multiple branches with atrous convolutions of different dilations were combined together to utilize multi-scale properties for image segmentation.
\paragraph*{Fully connected conditional random field}
Conditional random field is a undirected discriminative probabilistic graphical model that is often used for various sequence learning problems. Unlike discrete classifiers, while classifying a sample it takes into account the labels of other neighboring samples. Image segmentation can be treated as a sequence of pixel classifications. The label of a pixel is not only dependent on its own intensity values but also the values of neighboring pixels. The use of such probabilistic graphical models is often used in the field of image segmentation and hence it deserves a dedicated section (section \ref{sec:pgm}).

\subsubsection{Using inter pixel correlation to improve CNN based segmentation}
\label{sec:pgm}
The use of probabilistic graphical models such as markov random fields (MRF) or conditional random fields (CRF) for image segmentation thrived on its own even without the inclusion of CNN based feature extractors. The CRF or MRF is mainly characterized by an energy function with a unary and a pairwise component.
\begin{equation}
E(x) = \underbrace{\sum_i{\theta_i(x_i)}}_{\text{unary potential}} + \underbrace{\sum_{ij}{\theta_{ij}(x_i,x_j)}}_{\text{pairwise potential}}
\end{equation}
While non-deep learning approaches focused on building efficient pair-wise potentials like exploiting long-range dependencies, designing higher-order potentials and exploring contexts of semantic labels, deep learning based approaches focused on generating a strong unary potentials and using simple pairwise components to boost the performance. CRFs have usually been coupled with deep learning based methods in two ways. One as a separate post-processing module and the other as an trainable module in an end-to-end network like deep parsing networks\cite{deepparsing} or spatial propagation networks\cite{spatialpropagation}.
\paragraph{Using CRFs to improve Fully convolutional networks}
One of the earliest implementations that kick-started this paradigm of boundary refinements was the works of \cite{crf} With the introduction of fully convolutional networks for image segmentation it was quite possible to draw coarse segments for objects in images. However, getting sharper segments was still a problem. In the works of \cite{iclr_chen}, the output pixel level prediction was used as a unary potential for a fully connected CRF. For each pair of pixels $i$ and $j$ in the image the pairwise potential was defined as

\begin{multline}
\theta_{ij}(x_i,x_j) =\mu(x_i,x_j)\left[w_1\exp\left(-\frac{||p_i-p_j||^2}{2\sigma_\alpha^2} -\frac{||I_i-I_j||^2}{2\sigma_\beta^2}\right) \right. \\ \left. +w_2\exp\left(-\frac{||p_i-p_j||^2}{2\sigma_\gamma^2}\right)  \right]
\end{multline}

Here, $\mu(x_i,x_j) = 1 \text{ if } x_i\neq x_j, \text{0 otherwise}$ and $w_1, w_2$ are the weights given to the kernels. The expression uses two gaussian kernels. The first one is a bilateral kernel that depends on both pixel positions($p_i,p_j$) and their corresponding intensities in the RGB channels. The second kernel is only dependent on the the pixel positions. $\sigma_\alpha, \sigma_\beta \text{ and } \sigma_\gamma$ controls the scale of the Gaussian kernels. The intuition behind the design of such a pairwise potential energy function is to ensure that nearby pixels of similar intensities in the RGB channels are classified under the same class. This model has also been later included in the popular network called DeepLab (refer section \ref{sec:deeplab}). In the various versions of the DeepLab algorithm, the use of CRF was able to boost the mean IOU on the Pascal 2012 Dataset by significant amount(upto 4\% in some cases).

\paragraph{CRF as RNN}
While CRF is an useful post-processing module\cite{crf} for any deep learning based semantic image segmentation architecture, yet one of the main drawbacks was that it could not be used as a part of an end-to-end architecture. In the standard CRF model the pairwise potentials can be represented in terms of a sum of weighted Gaussians. However since the exact minimization is intractable a mean-field approximation of the CRF distribution is considered to represent the distribution with a simpler version which is simply a product of independent marginal distributions. This mean-field approximation in its native form isn't suitable for back-propagation. In the works of \cite{crfasrnn}, this step was replaced by a set of convolutional operation that is iterated over a recurrent pipeline until convergence is reached. As reported in their work, with the proposed approach an mIOU of 74.7 was obtained as compared to 71.0 by BoxSup and 72.7 by DeepLab.
The sequence of operations can be most easily explained as follows.

\begin{enumerate}
	\item Initialization : A SoftMax operations over the unary potentials can give us the intial distribution to work with.
	\item Message Passing : Convoluting using two Gaussian kernels, one spatial and one bilateral kernel. Similar to the actual implementation of CRF, the splatting and slicing also occurs while building the permutohedral lattice for efficient computation of the fully connected CRF
	\item Weighting Filter Outputs : Convoluting with $1 \times 1$ kernels with the required number of channels the filter outputs can be weighted and summed. The weights can be easily learnt through backpropagation.
	\item Compatibility Transform : Considering a compatibility function to keep a track of uncertainty between various labels, a simple $1\times 1$ convolution with the same number of input and output channel is enough to simulate that. Unlike the potts model that assigns the same penalty, here the compatibility function can be learnt and hence a much better alternative.
	\item Adding the unary potentials : This can be performed by a simple element wise subtraction of the penalty from the compatibility transform from the unary potentials
	\item Normalization : The outputs can be normalized with another simple softmax function. 
\end{enumerate}

\paragraph{Incorporating higher order dependencies} 
Another end-to-end network inspired from CRFs, incorporate higher order relations into a deep network . With a deep parsing network \cite{deepparsing} pixel-wise prediction from a standard VGG-like feature extractor (but with lesser pooling operations) is boosted using a sequence of special convolution and pooling operations. Firstly , by using local convolutions that implement large unshared convolutional kernels across the different positions of the feature map, to obtain translation dependent features that model long-distance dependencies. Similar to standard CRFs a spatial convolution penalizes probabilistic maps based on local label contexts. Finally, with block min pooling that does a pixel-wise min-pooling across the depth to accept the prediction with the lowest penalty. Similarly, in the works of \cite{spatialpropagation}, a row/columnwise propagation model was proposed the calculated the global pairwise relationship across an image. With a dense affinity matrix drawn from a sparse transformation matrix, coarsely predicted labels were reclassified based on the affinity of pixels.

\subsubsection{Multi-scale networks}
One of the main problems with image segmentation for natural scene images is that the size of the object of interest is very unpredictable, as in real world objects may be of different sizes and objects may look bigger or smaller depending on the position of the object and the camera. The nature of a CNN dictates that delicate small scale features are captured in early layers whereas as one moves across the depth of the network the features become more specific for larger objects. For example a tiny car in a scene has much lesser chance of being captured in the higher layers due to operations like pooling or down-sampling. It is often beneficial to extract information from feature maps of various scales to create segmentations that are agnostic of the size of the object in the image. Multi-scale auto-encoder models ~\cite{seg_attention2scale} consider activations of different resolutions to provide image segmentation output.

\begin{figure}[htbp]
	\centering
	\includegraphics[width=0.9\textwidth]{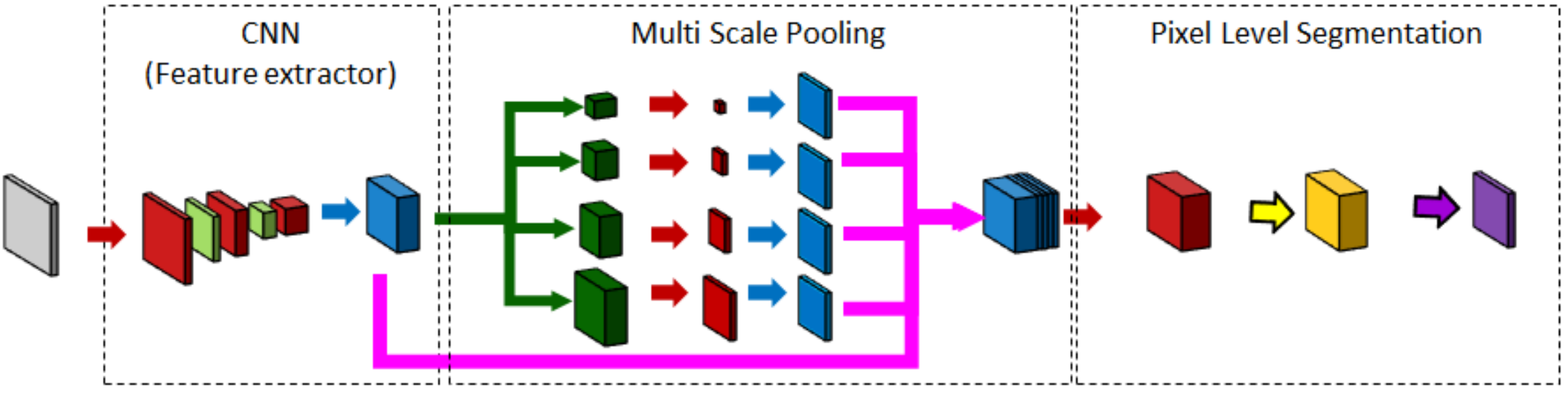}
	\caption{A schematic representation of the PSPNet}
	\label{fig:pspnet}
\end{figure}

\paragraph*{PSPNet}
The pyramid scene parsing network~\cite{seg_pspnet} was built upon the FCN based pixel level classification network. The feature maps from a ResNet-101 network are converted to activations of different resolutions thorough multi-scale pooling layers which are later upsampled and concatenated with the orginal feature map to perform segmentation(Refer fig.\ref{fig:pspnet}). The learning process in deep networks like ResNet was further optimized by using auxiliary classifiers. The different types of pooling modules focus on different areas of the activation map. Pooling kernels of various sizes like $1\times 1$, $2\times 2$, $3\times 3$, $6\times 6$ look into different areas of the activation map to create the spatial pooling pyramid. One the ImageNet scene parsing challenge the PSPNet was able to score an mean IoU of 57.21 with respect to 44.80 of FCN and 40.79 of SegNet.

\begin{figure}[htbp]
	\centering
	\includegraphics[width=0.9\textwidth]{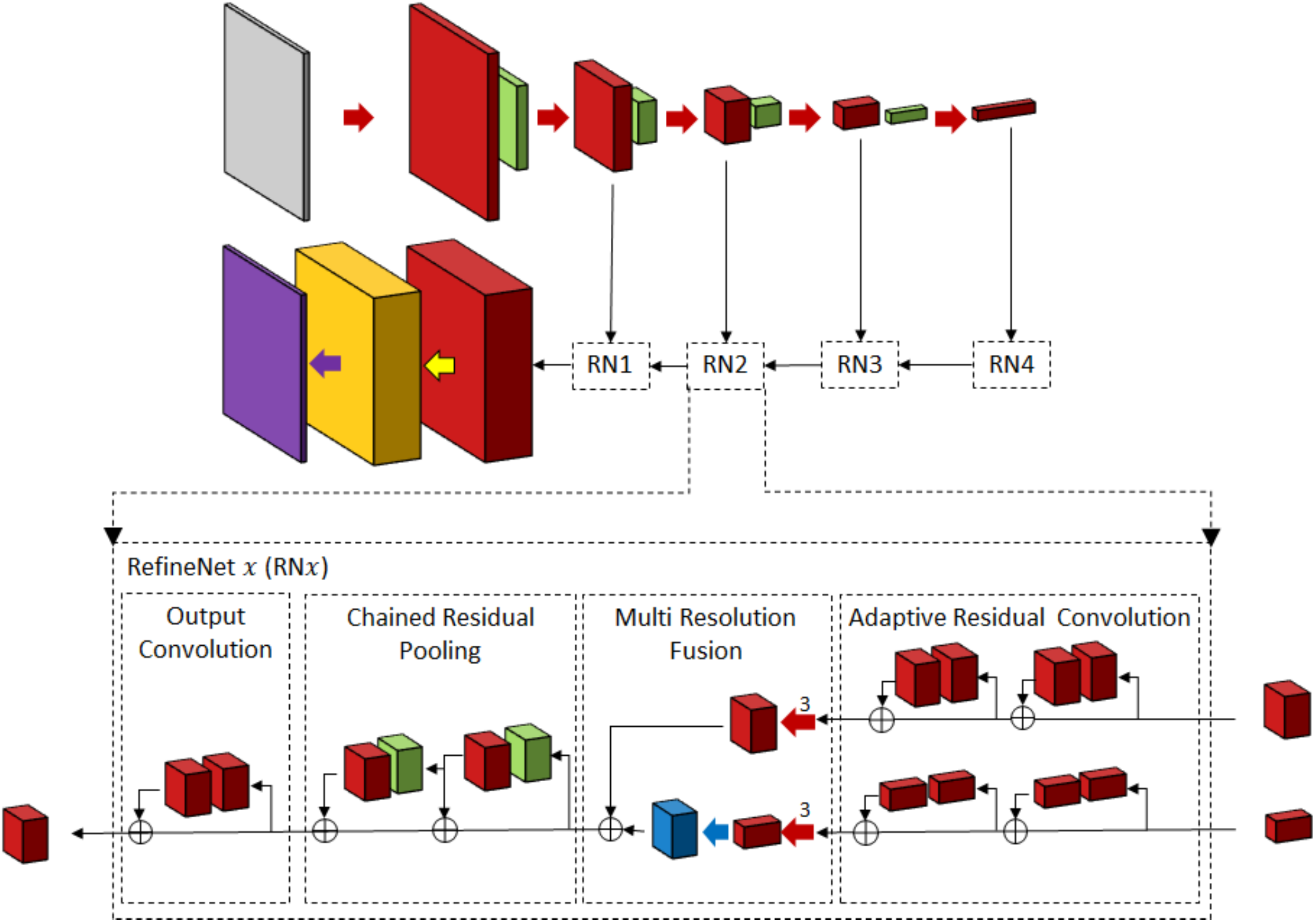}
	\caption{A schematic representation of the RefineNet}
	\label{fig:refinenet}
\end{figure}

\paragraph*{RefineNet}
Working with features from last layer of a CNN produces soft boundaries for the object segments. This issue was avoided in DeepLab algorithms with atrous convolutions. RefineNet~\cite{seg_refinenet} takes an alternative approach by refining intermediate activation maps and hierarchically concatenating it to combine multi-scale activations and prevent loss of sharpness simultaneously. The network consisted of separate RefineNet modules for each block of the ResNet. Each RefineNet module were made up of three main blocks, namely, Residual convolution unit(RCU), multi-resolution fusion(MRF) and chained residual pooling(CRP)(Refer fig.\ref{fig:refinenet}). The RCU block consists of an adaptive convolution set that fine-tunes the pre-trained weights of the ResNet weights for the segmentation problem. The MRF layer fuses activations of different resolutions using convolutions and upsampling layers to create a higher resolution map. Finally in CRP layer pooling kernels of multiple sizes are used on the activations to capture background context from large image areas. The RefineNet was tested on the Person-Part Dataset where it obtained an IOU of 68.6 as compared to 64.9 by DeepLab-v2 both of which used the ResNet-101 as a feature extractor. 

\subsection{Convolutional autoencoders}
The last subsection deals with discriminative models that are used to perform pixel level classification to deal with image segmentation problems. Another line of thought gets its inspiration from autoencoders. Autoencoders have been traditionally used for feature extraction from input samples while trying to retain most of the original information. An autoencoder is basically composed of an encoder that encodes the input representations from a raw input to a possibly lower dimensional intermediate representation and a decoder that attempts to reconstruct the original input from the intermediate representation. The loss is computed in terms of the difference between the raw input images and the reconstructed output image. The generative nature of the decoder part has often been modified and used for image segmentation purposes. Unlike the traditional autoencoders, during segmentation the loss is computed in terms of the difference between the reconstructed pixel level class distribution and the desired pixel level class distribution. This kind of segmentation approach is more of a generative procedure as compared to the classification approach of RCNN or DeepLab algorithms. The problem with approaches such as this is to prevent over-abstraction of images during the encoding process. The primary benefit of such approaches is the ability to generate sharper boundaries with much lesser complication. Unlike the classification approaches, the generative nature of the decoder can learn to create delicate boundaries based on extracted features. The major issue that affects these algorithm is the level of abstraction. It has been seen that without proper modification the reduction in the size of the feature map created inconsistencies during the reconstruction. in the paradigm of convolutional neural networks the encoding is basically a series of convolution and pooling layers or strided convolutions. The reconstruction however can be tricky. The commonly used techniques for decoding from a lower dimensional feature are transposed convolution or a unpooling layers. One of the main advantages of using autoencoder based approach over normal convolutional feature extractor is the freedom of choosing input size. With a clever use of down-sampling and up-sampling operation it is possible to output a pixel-level probability that is of the same resolution as the input image. This benefit has made encoder-decoder architectures with multi-scale feature forwarding has become ubiquitous for networks where input size is not predetermined and an output of same size as the input is needed. 

\begin{figure}[htbp]
	\centering
	\includegraphics[width=0.6\textwidth]{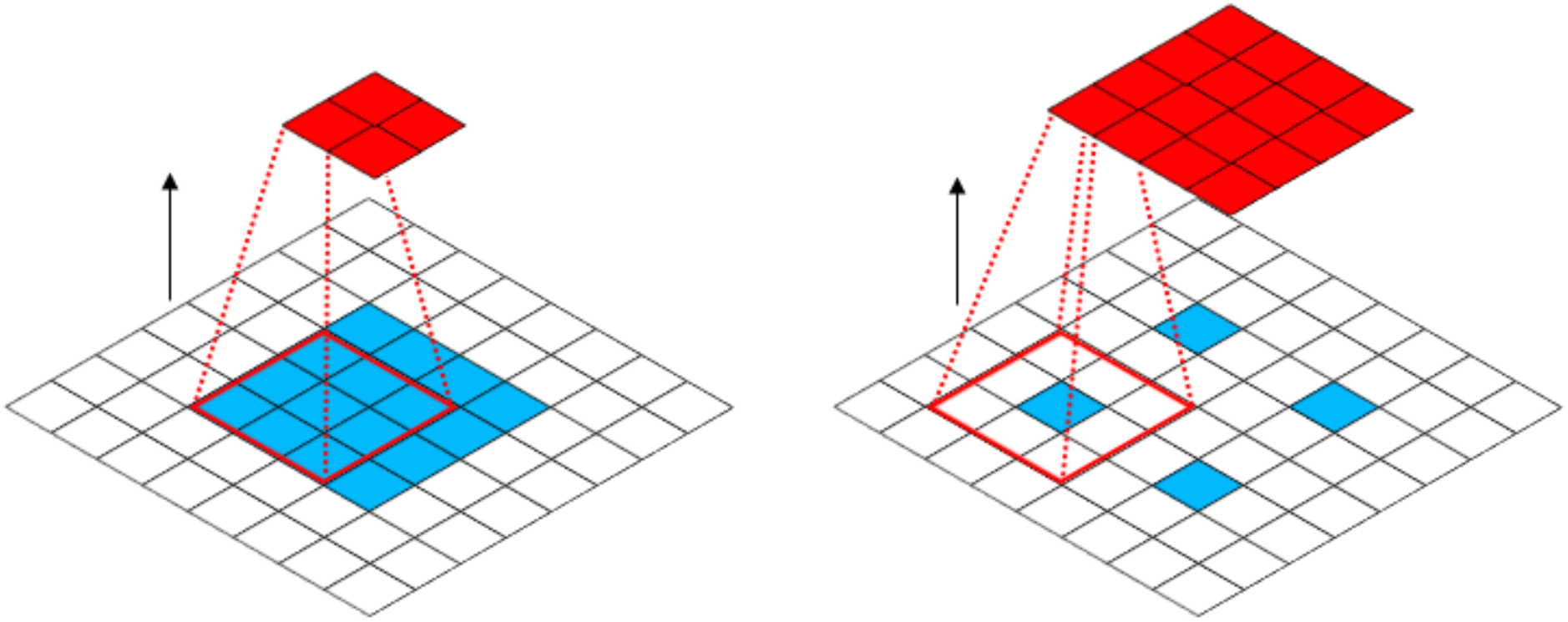}
	\caption{(Left) Normal Convolution with unit stride. (Right) Transposed convolution with fractional strides.}
	\label{fig:transposed}
\end{figure}

\paragraph*{Transposed Convolution} Transposed convolution also known as convolution with fractional strides has been introduced to reverse the effects of a traditional convolution operation ~\cite{met_deconv,met_deconv2}. It is often referred to as deconvolution. However deconvolution, as defined in signal processing, is different than transposed convolution in terms of the basic formulation, although they effectively address the same problem. In a convolution operation there is a change in size of the input based on the amount of padding and stride of the kernels. As shown in fig. \ref{fig:transposed} a stride of 2 will create half the number of activations as that of a stride of 1. For a transposed convolution to work padding and stride should be controlled in a way that the size change is reversed. This is achieved by dilating the input space. Note that unlike atrous convolutions, where the kernels were dilated, here the input spaces are dilated.
\paragraph*{Unpooling} Another approach to reduce the size of the activations is through pooling layers. a $2\times 2$ pooling layer with a stride of two reduces the height and width of the image by a factor of 2. In such a pooling layer, a $2\times2$ neighborhood of pixel is compressed to a single pixel. Different types of pooling performs the compression in different ways. Max-pooling considers the maximum activation value among 4 pixels while average pooling takes an average of the same. A corresponding unpooling layer decompresses a single pixel to a neighborhood of $2\times2$ pixels to double the height and width of the image.	
\subsubsection{Skip Connections} 
\begin{figure}[htbp]
	\centering
	\includegraphics[width=0.75\textwidth]{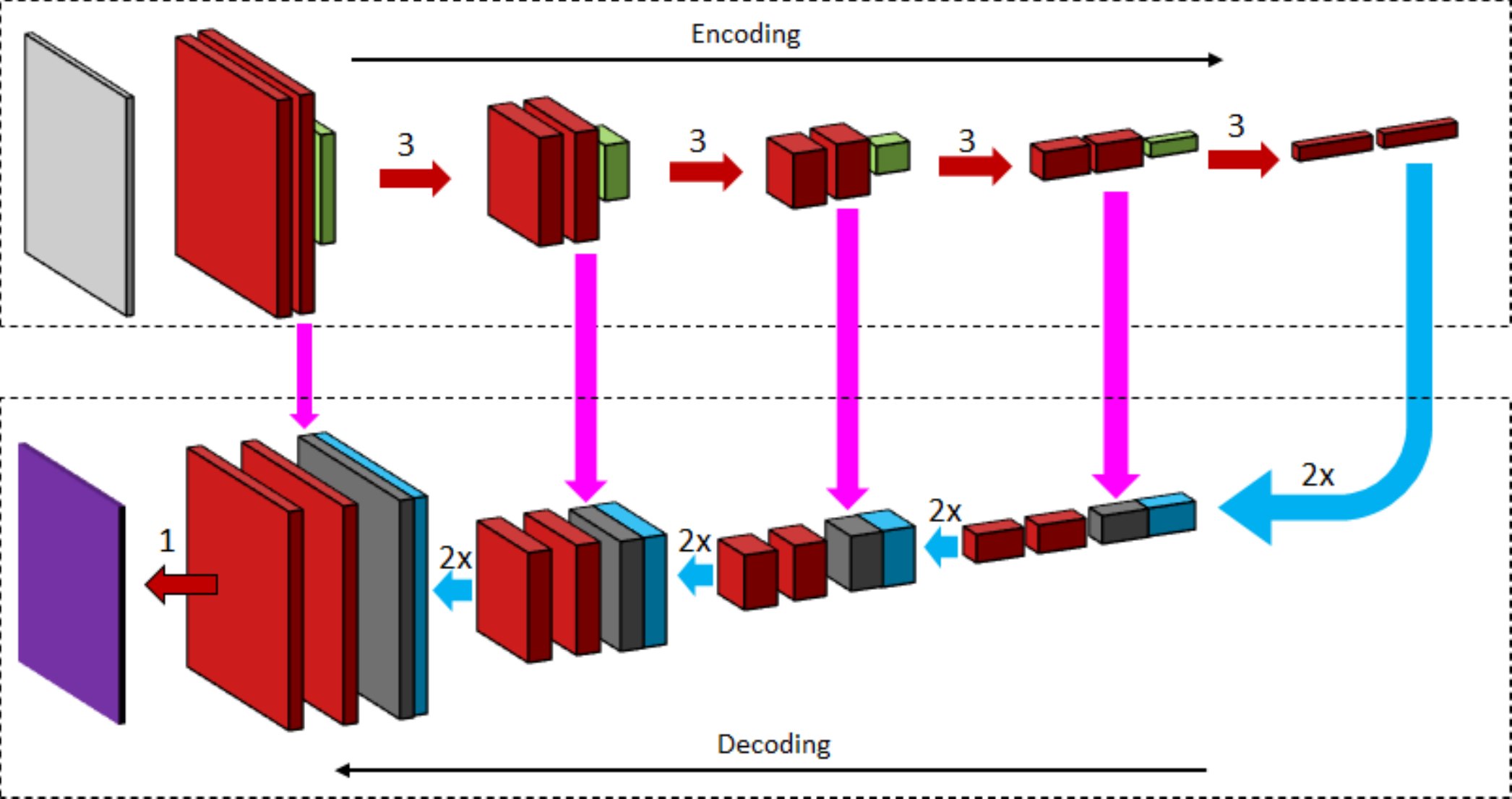}
	\caption{Architecture of U-Net}
	\label{fig:unet}
\end{figure}

Linear skip connections has often been used in convolutional neural networks to improve gradient flow across a large number of layers~\cite{deep_resnet}. As depth increases in a network the activations maps tend to focus on more and more abstract concepts. Skip connections has proved to be very effective to combine different levels of abstractions from different layers to generate crisp segmentation maps. 
\paragraph*{U-NET} The U-Net architecture, proposed in 2015, proved to be quite efficient for a variety of problems such as segmentation of neuronal structures, radiography, and cell tracking challenges ~\cite{seg_unet}. The network is characterized by an encoder with a series of convolution and max pooling layers. The decoding layer contains a mirrored sequence of convolutions and transposed convolutions. As described till now it behaves as a traditional auto-encoder. Previously it has been mentioned how the level of abstraction plays an important role in the quality of image segmentation. To consider various levels of abstraction U-Net implements skip connections to copy the uncompressed activations from encoding blocks to their mirrored counterparts among the decoding blocks as shown in the fig. \ref{fig:unet}. The feature extractor of the U-Net can also be upgraded to provide better segmentation maps. The network nicknamed "The one hundred layers Tiramisu" ~\cite{seg_tiramisu} applied the concept of U-Net using a dense-net based feature extractor. Other modern variations involve the use of capsule networks~\cite{dynamicrouting} along with locally constrained routing~\cite{seg_caps}. U-Net was selected as a winner for an ISBI cell tracking challenge. In the PhC-U373 dataset it scored a mean IoU of 0.9203 whereas the second best was at 0.83. In the DIC-HeLa dataset, it scored a mean IoU of 0.7756 which was significantly better than the second best approach which scored only 0.46.

\subsubsection{Forwarding pooling indices} 
\begin{figure}[htbp]
	\centering
	\includegraphics[width=0.6\textwidth]{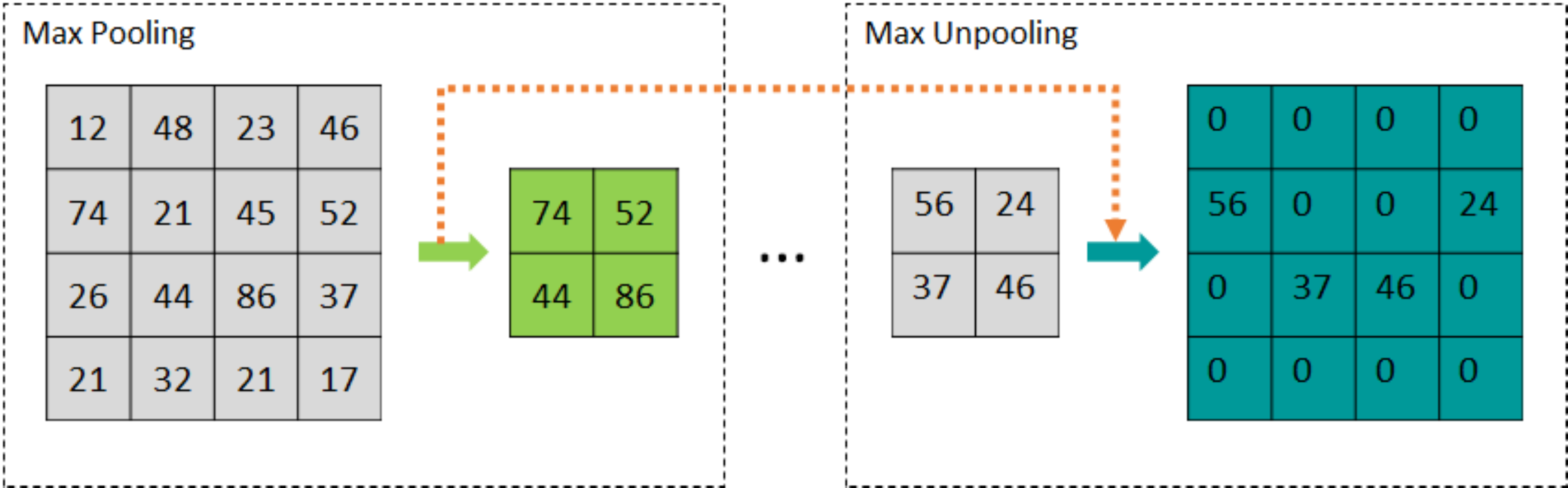}
	\caption{Forwarding pooling indices to maintain spatial relationship during unpooling.}
	\label{fig:pooling}
\end{figure}

Max-pooling has been the most commonly used technique for reducing the size of the activation maps for various reasons. The activations represent of the response of the region of an image to a specific kernel. In max pooling, a region of pixels is compressed to single value by considering only the maximum response obtained within that region. If a typical autoencoder compresses a $2\times 2$ neighborhood of pixels to a single pixel in the encoding phase, the decoder must decompress the pixel to a similar dimension of $2\times 2$. By forwarding pooling indices the network basically remembers the location of the maximum value among the 4 pixels while performing max-pooling. The index corresponding to the maximum value is forwarded to the decoder(Refer fig.\ref{fig:pooling}) so that while the un-pooling operation the value from the single pixel can be copied to the corresponding location in $2\times 2$ region in the next layer ~\cite{met_unpooling}. The values in rest of the three positions are computed in the subsequent convolutional layers. If the value was copied to random location without the knowledge of the pooling indices, there would be inconsistencies in classification especially in the boundary regions. 

\begin{figure}[htbp]
	\centering
	\includegraphics[width=0.75\textwidth]{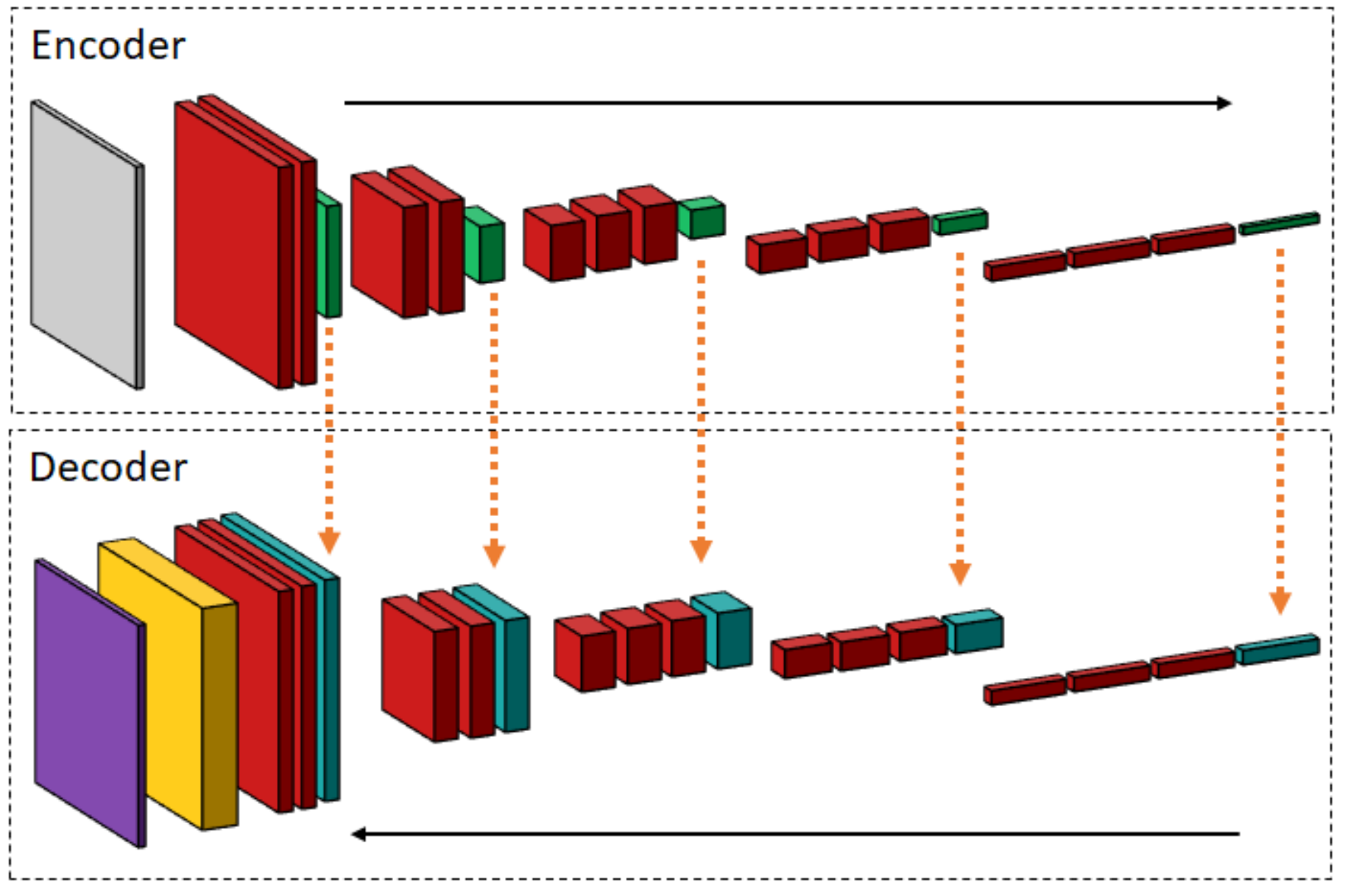}
	\caption{Architecture of SegNet}
	\label{fig:segnet}
\end{figure}

\paragraph*{SegNet}
The SegNet algorithm~\cite{seg_segnet} was launched in 2015 to compete with the FCN network on complex indoor and outdoor images. The architecture was composed of 5 encoding blocks and and 5 decoding blocks. The encoding blocks followed the architecture of the feature extractor in VGG-16 network. Each block is a sequence of multiple convolution, batch normalization and ReLU layers. Each encoding block ends with a max-pooling layer where the indices are stored. Each decoding block begins with a unpooling layer where the saved pooling indices are used (Refer fig.\ref{fig:segnet}). The indices from the max-pooling layer of the  $ith$ block in the encoder is forwarded to the max-unpooling layer in the  $(L-i+1)th$ block in the decoder where $L$ is the total number of blocks in each of the encoder and decoder. The SegNet architecture scored an mIoU of 60.10 as compared to 53.88 by DeepLab-LargeFOV\cite{deep_deeplab} or 49.83 by FCN\cite{seg_fcn} or 59.77 by Deconvnet\cite{met_deconv} on the CamVid Dataset.

\subsection{Adversarial Models}
Until now, we have seen purely discriminative models like FCN, DeepMask, DeepLab that primarily generates a probability distribution for every pixel across the number of classes. Furthermore, autoencoder treated segmentation as a generative process however the last layer is generally connected to a pixel-wise soft-max classifier. The adversarial learning framework approaches the optimization problem from a different perspective. Generative Adversarial Networks (GANs) gained a lot of popularity due to there remarkable performance as a generative network. The adversarial learning framework mainly consists of two networks a generative network and a discriminator network. The generator $G$ tries to generate images,,./ like the ones from the training dataset using a noisy input prior distribution called $p_{z}(z)$. The network $G(z;\theta_g)$ represents a differentiable function represented by a neural network with weights $\theta_g$. A discriminator network tries to correctly guess whether an input data is from the training data distribution ($p_{data}(x)$) or generated by the generator $G$. The goal of the discriminator is to get better at catching a fake image, while the generator tries to get better at fooling the discriminator, thus in the process generating better outputs. The entire optimization process can be written as a min-max problem as follows:
\begin{multline}
\min_G \max_D V(D,G) = \mathbb{E}_{x\sim p_{data}(x)}[\log D(x)] +\\ \mathbb{E}_{z\sim p_z(z)}[\log(1-D(G(z)))]
\end{multline}
The segmentation problem has also been approached from a adversarial learning perspective. The segmentation network is treated as a generator that generates the segmenation masks for each class, whereas a discriminator network tries to predict whether a set of masks is from the ground truth or from the output of the generator~\cite{seg_gan}. A schematic diagram of the process is shown in fig.\ref{fig:gan}. Furthermore, conditional GANs have been used to perform image to image translation\cite{seg_cgan}. This framework can be used for image segmentation problems where the semantic boundaries of the image and output segmentation map do not necessarily coincide, for example, in case of creating a schematic diagram of a fa\c{c}ade of a building.
\begin{figure}[htbp]
	\centering
	\includegraphics[width=0.75\textwidth]{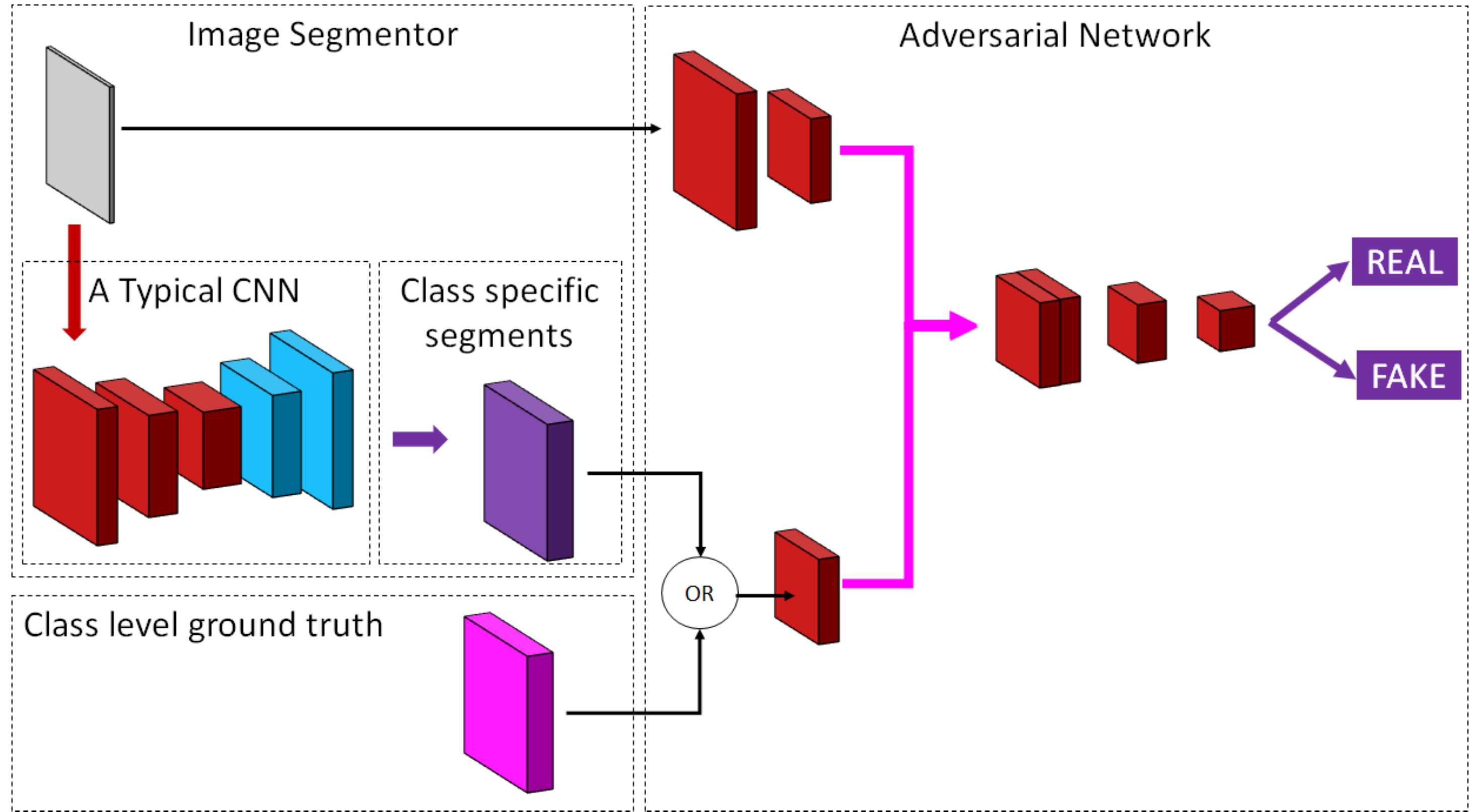}
	\caption{Adversarial learning model for image segmentation}
	\label{fig:advseg}
\end{figure}
\subsection{Sequential Models}
Till now almost all the techniques discussed deal with semantic image segmentation. Another class of segmentation problem, namely, instance level segmentation needs slightly different approach. Unlike semantic image segmentation, here all instances of the same object are segmented into different classes. This type of segmentation problem is mostly handled as a learning to give a sequence of object segments as outputs. Hence sequential models come into play in such problems. Some of the main architectures commonly used are convolutional LSTMs, Reccurent Networks, Attention-based models and so on. 
\subsubsection{Recurrent Models}
Traditional LSTM networks employ fully connected weights to model long and short term memories accross sequential inputs. But they fail to capture spatial information of images. Moreover, fully connected weights for images increases the cost of computation by a great extent. In convolutional LSTM~\cite{seg_rnn} these weights are replaced by convolutional layers (Refer fig. \ref{fig:lstm}). Convolutional LSTMs have been used in several works to perform instance level segmentation. Normally they are used as a suffix to a object segmentation network. The purpose of the recurrent model like LSTM is to select each instance of the object in different timestamps of the sequential output. The approach has been implemented with object segmentation frameworks like FCN and U-NET~\cite{seg_rnn_unet}. 
\subsubsection{Attention Models}
While convolutional LSTMs can select different instance of objects at different timestamps, attention models are designed to have more control over this process of localizing individual instances. One simple method to control attention is by spatial inhibition~\cite{seg_rnn}. Spatial inhibition network is designed to learn a bias parameter that cuts off previously detected segments from future activations. Attention models have been further developed with the introduction of dedicated attention module and an external memory to keep track of segments. In the works of ~\cite{seg_attention}, the instance segmentation network was divided into 4 modules. First, and external memory provides object boundary details from all previous steps. Second, a box network attempts to predict the location of the next instance of the object and outputs a sub-region of the image for the third module that is the segmentation module. The segmentation module is similar to a convolutional auto-encoder model discussed previously. The fourth module scores the predicted segments based on whether they qualify as a proper instance of the object. The network terminates when the score goes below a user-defined threshold.
\subsection{Weakly Supervised or Unsupervised Models}
Neural Networks in general are trained with algorithms like back-propagation, where the parameters $\textbf{w}$ are updated based on their local partial derivative with respect to a error value $E$ obtained using a loss function $f$.
\begin{equation}
w = w + \Delta w = w -\eta \frac{\delta E}{\delta w}
\end{equation}
The loss function is generally expressed in terms of a distance between a target value and the predicted value. But in many scenarios image segmentation requires the use of data without annotations with ground truth. This leads to the development of unsupervised image segmentation techniques. One of the straight forward ways to achieve this is to use networks pre-trained on other larger datasets with similar kinds of samples and ground truths and use clustering algorithms like K-means on the feature maps. However this kind of semi-supervised technique is inefficient for data samples that have a unique distribution of sample space. Another cons is that the network is trained to perform on a input distribution which is still different from the test data. That does not allow the network to perform to it with full potential. The key problem in fully unsupervised segmentation algorithm is the development of a loss function capable of measuring the quality of segments or cluster of pixels. With all these limitations the amount of literature is comparatively much lighter when it comes to weakly supervised or unsupervised approaches.
\subsubsection{Weakly supervised algorithms}
Even in the lack of proper pixel level annotations, segmentation algorithms can exploit coarser annotations like bounding boxes or even image level labels\cite{weaklybox,li2017not} for performing pixel level segmentation.
\paragraph*{Exploiting bounding boxes}
From the angle of data annotation, defining bounding boxes is a much less expensive task as compared to pixel level segmentation. The availability of datasets with bounding boxes is also much larger than those with pixel level segmentations. The bounding box can be used as a weak supervision to generate pixel level segmentation maps. In the works of \cite{dai2015boxsup}, titled BoxSup, segmentation proposals were generated using region proposal methods like selective search. After that multi-scale combinatorial grouping is used to combine candidate masks and the objective is to select the optimal combination that has the highest IOU with the box. This segmentation map is used to tune a traditional image segmentation network like FCN. BoxSup was able to attain an mIOU of 75.1 in the pascal VOC 2012 test set as compared to 62.2 of FCN or 66.4 of DeepLab-CRF.
\subsubsection{Unsupervised Segmentation}
Unlike supervised or weakly supervised segmentation, the success of unsupervised image segmentation algorithms are mostly dependent on the learning mechanism. Some of the common approaches are described below. 
\paragraph*{Learning multiple of objectives}
One of the most generic approach in unsupervised learning is considering multiple objectives designed in a way to perform well for segmentation when ground-truths are not available.
\par A common variant called JULE or joint unsupervised learning of deep representation have been used in several applications where there is a lack of samples with ground-truth. The basis of JULE lies in training a sequential model along with a deep feature extraction model. The learning methodology was primarily introduced as an image clustering algorithm. However it has been extended to other applications like image segmentation. The key objective for being able to perform these kinds of segmentation is the development of a proper objective function. In JULE, the objective function considers the affinity between samples in the clusters and also the negative affinity between the clusters and its neighbors. Agglomerative clustering is performed across the timestamps of a recurrent network. In the works of \cite{unsup1}, JULE was attempted on image patches rather than entire samples of images. With a similar objective function, it was able to classify the patches into a predefined number of classes. JULE was used in this case to provide the agglomerative clustering information as supervisory signals for the next iteration.
\par Another variant of learning multiple objectives has been demonstrated through adversarial collaboration\cite{adversarialcollaboration} among neural networks with independent jobs like monocular depth prediction, estimating camera motion, detecting optical flow and segmentation of a video into the static scene and moving regions. Through competitive collaboration, each network competes to explain the same pixel that either belongs to static or moving class which in turn shares their learned concepts with a moderator to perform the motion segmentation.
\paragraph*{Using refinement modules for self supervision}:
Using other unsupervised over-segmentation techniques can be used to provide supervision to deep feature extractors \cite{kanezaki2018unsupervised}. By enforcing multiple constraints like similarity between features, spatial continuity, intra-axis normalization. All these objectives are optimized through back propagation. The spatial continuity is achieved by extracting superpixel from the image using standard algorithms like the SLIC\cite{slic} and all pixels within a superpixel are forced to have the same label. The difference between the two segmentation map is used as a supervisory signal to update the weights.
\paragraph{Other relevant unsupervised techniques for extracting semantic information:}
Learning without annotations is always challenging and hence have a variety of literature where many interesting solutions have been proposed. Using CNNs to solve jigsaw puzzles ~\cite{new1} derived from images can be used to learn semantic connections between various parts of the objects. The proposed context free network takes a set of image tiles as input and tries to establish the correct spatial relations between them. During the process it simultaneously learns features specific to parts of an object as well as their semantic relationships. These patch based self supervision techniques can be further improved using contextual information \cite{new2}. Using context-encoders \cite{new3} can also derive spatial and semantic relationship among various parts of an image. Context encoders are basically, CNNs trained to generate arbitrary regions of an image which is conditioned by its surrounding information. Another demonstration of extracting semantic information can be found in the process of image colorization \cite{new4}. The process of colorization requires pixel level understanding of the semantic boundaries corresponding to objects. Other self supervision techniques can leverage motion cues in videos to segment various parts of an object for better semantic understanding \cite{new5}. This method can learn several structural and coherent features for tasks like semantic segmentation, human parsing, instance segmentation and so on.

\subsubsection{W-Net}
W-Net~\cite{seg_Wnet} derived its inspiration from the previously discussed U-Net. The W-Net architecture consists of a two cascaded U-Nets. The first U-Net acts as a encoder that converts an image to its segmented version while the second U-Net tries to reconstruct the original image from the output of the first U-Net that is the segmented image. Two loss functions are minimized simultaneously. One of them being the Mean square error between the input image and the reconstructed image that is given by the second U-Net. The second loss function is derived from the Normalized-Cut~\cite{met20_normalized}. The hard normalized cut is formulated as,
\begin{equation}
N cut_K(V) = \sum_{k=1}^K \frac{\sum_{u\in A_k, v\in V-A_k}w(u,v)}{\sum_{u\in A_k, t\in V}w(u,t)}
\end{equation}
where, $A_k$ is set of pixels in the $k-th$ segment, $V$ is the set of all the pixels, and $w$ measures the weight between two pixels.
\par However, this function is not differentiable and hence backpropagation is not possible. Hence a soft version of function was proposed.
\begin{equation}
J_{soft-Ncut}(V,K) = K - \sum_{k=1}^K \frac{\sum_{u\in V}p(u=A_k)\sum_{v\in V}w(u,v)p(v=A_k)}{\sum_{u\in V}p(u=A_k)\sum_{t\in V}w(u,t)}
\end{equation}
where, $p(u=A_k)$ represents the probability of a node $u$ belonging to a class $A_k$. The output segmentation maps were further refined using fully connected conditional random fields. Remaining insignificant segments was further merged using hierarchical clustering.
\subsection{Interactive Segmentation}
Image segmentation is one of the most difficult challenges in the field of computer vision. In many scenarios where the images are too complex, noisy or subjected to poor illumination conditions, a little bit of interaction and guidance from users can significantly improve the performance of segmentation algorithms. Interactive segmentation have been flourishing even outside the deep learning paradigm. However with powerful feature extraction of convolutional neural nets the amount of interaction can be reduced to get extent.
\subsubsection{Two stream fusion}
One of the most straight forward implementations of interactive segmentation is to have two parallel branches, one from the image another from the an image representing the interactive stream, and fuse them to perform the segmentation\cite{twostreamfusion}. The interaction inputs are taken in form of different coloured dots representing positive and negative classes. With a bit of post processing where intensities of the interaction maps are calculated based on the euclidean distance from the points we get two sets of maps(one for each class) that looks like fuzzy voronoi cells with the points at the centre. The maps are multiplied element-wise to obtain the image for the interaction stream. The two branches are composed of sequence of convolutions and pooling layers. The Hadamard product of features obtained at the end of each branches are sent to the fusion network where a low resolution segmentation map is generated. An alternative approach follows the footsteps of FCN to fuse features of multiple scales to obtain a sharper resolution image.
\subsubsection{Deep Extreme Cut}
Contrary to the two stream approach, deep extreme cut\cite{deepextremecut} takes a single pipeline to create segmentation maps from RGB Images. This method expects 4 points from the user denoting the four extreme regions in the boundary of the object (leftmost, rightmost, topmost,bottommost). By creating heatmaps from the points, a 4 channel input is fed into a DenseNet101 network. The final feature map of the network is passed into a pyramid scene parsing module for analyzing global contexts to perform the final segmentation. This method was able to attain an mIOU of 80.3 on the PASCAL test set.
\subsubsection{Polygon-RNN}
Polygon-RNN\cite{polygonrnn} takes a different approach to the other methods. Multi scale features are extracted from different layers of a typical VGG Network and concatenated to create a feature block for a recurrent network. The RNN in turn is supposed to provide a sequence of points as an output that represents the contour of the object. The system is primarily designed as an interactive image annotation tool. The users can interact in two different ways. Firstly the users must provide a tight bounding box for the object of interest. Secondly after the polygon is built the users were allowed to edit any point in the polygon. However this editing is not used for any further training of the system and hence presents a small avenue for improvement of the system.
\subsection{Building more efficient networks}
While many complicated networks with lots of fancy modules can give a very decent quality of semantic segmentation, embedding such algorithms in real-life systems is a different story. Many other factors like cost of hardware, real-time response and so on poses a new degree of challenge. Efficiency is also key for creating consumer level systems. 
\subsubsection{ENet}
The ENet\cite{enet} brought forward a couple of interesting design choices to create a quite shallow segmentation network with a small number of parameters (0.37 Million). Instead of a symmetric encoder decoder architecture like SegNet or U-Net, it has a deeper encoder and a shallower decoder. Instead of increasing channel sizes after pooling, parallel pooling operations were performed along with convolutions of stride 2 to reduce overall features. To increase the learning capability PReLU were used as compared to ReLU so that the transfer functions remains dynamic so that it can simulate the jobs of a ReLU as well as an identity function as required. This is normally an important factor in ResNet however because the network is shallow, using PReLU is a smarter choice. Above that using factorized filters also allowed for a smaller number of parameters.
\subsubsection{Deep Layer Cascade}
Deep Layer Cascade \cite{li2017not} tackles several challenges and makes two significant contributions. Firstly, it analyzed the level of difficulty of pixel level segmentation for various classes. With a cascaded network, easier segments are discovered in the earlier stage while the latter layers focus on regions that need more delicate segments. Secondly, the proposed layer cascading can be used with common networks like Inception-ResNet-V2(IRNet) to improve the speed and even the performance to some extent. The basic principle of IRNet is to create a multi-stage pipeline where in each stage a certain amount of pixels would be classified into one of the segments. In the earlier stages the easiest pixels will be classified and the harder pixels with more uncertainty will move forward to latter stages. In the consequent stages the convolutions will only take place on those pixels which could not be classified in the previous stage while forwarding yet harder pixels to the next stage. Typically the proposed model comes with three stages each adding more convolutional modules to the network. With layer cascading an mIOU of 82.7 was reached on the VOC12 test set with DeepLabV2 and Deep Parsing Network being the nearest competitors with mIOU of 79.7 and 77.5 respectively. In terms of speed, 23.6 frames were processed per second as compared to 14.6 fps by SegNet or 7.1 fps by DeepLab-V2.
\subsubsection{SegFast}
Another recent implementation titled SegFast \cite{segfast} was able to build a network with only 0.6 Million parameters that resulted in a network that can do a forward pass in around 0.38 seconds without a GPU. The approach combined the concept of depth-wise separable convolutions with the fire modules of SqueezeNet. SqueezeNet introduced the concepts of fire modules to reduce the number of convolutional weights. With depth-wise separable convolutions, the number of parameters went further down. They also proposed the use of depthwise separable transposed convolutions for decoding. Even with so many approaches of feature reductions the performance was quite comparable to other popular networks like SegNet.
\subsubsection{Segmentation using superpixels}
Over-segmentation algorithms ~\cite{slic} have flourished well to divide images into small patches based on local information. With patch classification algorithms these superpixels can be converted to semantic segments. However since the process of over-segmentation do not consider neighborhood relations, it is necessary to include that in the patch classification algorithm. It is much faster to perform patch classification as compared to pixel level classification simply because of the fact that number of superpixels is much lesser than the number of pixels in an image. One of the first works of using CNNs for superpixel level classification was carried out by Farabet et al. ~\cite{farabet2013learning}. However just considering superpixels without context can result in errorneous classification. In the works of ~\cite{das2019combining}, multiple levels of contexts were captured by considering neighborhood super-pixels of different levels during the patch classification. By fusing patch level probabilities from different levels of contexts using methods like weighted averaging, max-voting or uncertainty based approaches like dempster-shafer theory, a very efficient segmentation algorithm was proposed. The works of Das et al. was able to obtain a pixel level accuracy of 77.14\%~\cite{das2019combining} with respect to 74.56\% Farabet et al. ~\cite{farabet2013learning}. While superpixels can be exploited to build efficient semantic segmentation models, the reverse is also true. Conversely, semantic segmentation ground-truths can be used to train networks to perform over-segmentation~\cite{tu2018learning}. Pixel affinities can be calculated using convolutional features to perform over-segmentation while paying special attention to semantic boundaries.
\section{Applications}
Image segmentation is one of the most commonly addressed problems in the domain of computer vision.  It is often augmented with other related tasks like object detection, object recognition, scene parsing, image description generation. Hence this branch of study finds extensive use in various real-life scenarios.
\subsection{Content-based image retrieval (CBIR)} 
With the ever increasing amount of structured and unstructured data on the internet, development of efficient information retrieval systems is of the utmost importance. CBIR systems have hence been a lucrative area of research. Many other related problems like visual question answering, interactive query based image processing, description generation. Image segmentation is useful in many cases as they are representative of spatial relations among various objects~\cite{app_cbir,app_cbir2}. Instance level segmentation is essential for handling numeric queries~\cite{app_cbir_inst}. Unsupervised approaches~\cite{app_cbir_unsup} are particularly useful for handling bulk amount of non-annotated data which is very common in this field of work.
\subsection{Medical imaging}
Another major application area for image segmentation is in the domain of health care. Many kinds of diagnostic procedures involve working with images corresponding to different types of imaging source and various parts of the body. Some of the most common types of tasks are segmentation of organic elements like, vessels~\cite{app_vessel}, tissues~\cite{app_tissue}, nerves~\cite{app_nerve}, and so on. Other kinds of problems include localization of abnormalities like tumors~\cite{app_tumor,app2_tumor}, aneurysms~\cite{app1_aneurysm,app_aneurysm2} and so on. Microscopic images~\cite{app_histo} also need various kinds of segmentations like cell or nuclei detection, counting number of cells, cell structure analysis for cancer detection and so on. The primary challenges with this domain is the lack of bulk amount of data for challenging diseases, variety in the quality of images due to the different types of imaging device involved. Medical procedures are not only involved to human beings, but also other animals as well as plants.
\subsection{Object Detection}
With the success of deep learning algorithms there has also been a surge in research areas related to automatic object detection. Many application like robotic maneuverability~\cite{app_robotics}, autonomous driving~\cite{app_driving}, intelligent motion detection~\cite{app_motion}, tracking systems~\cite{app_tracking2} and so on. Extremely remote regions such as deep sea~\cite{app_sea1,app_sea2}, or space~\cite{app_space} can be efficiently explored with the help of  intelligent robots making autonomous decisions. In sectors like defense, unmanned aerial vehicles or UAVs~\cite{app_uav} are used to detect anomalies or threats in remote regions~\cite{app9_military}. Segmentation algorithms have significant usage in satellite images for various geo-statistical analysis~\cite{app_satellite}. In fields like image or video post-production it is often essential to perform segmentation for various tasks like image matting\cite{matting,matting}, compositing\cite{compositing} and rotoscoping\cite{rotoscope}.
\subsection{Forensics}
Biometric verification systems like iris~\cite{app_iris1,app_iris2}, fingerprint~\cite{app_fingerprint}, finger vein~\cite{app_fingervein}, dental records~\cite{app_dental}, involve segmentation of various informative regions for efficient analysis.
\subsection{Surveillance}
Surveillance systems~\cite{app_news,app_crowd,app_action} are associated with various issues like occlusion, lighting or weather conditions. Moreover surveillance system can also involve analysis of images from hyper-spectral sources~\cite{app_hyperspectral}. Surveillance system can also be extended to various applications such as object tracking~\cite{app_tracking}, searching~\cite{app_searching}, anomaly detection~\cite{app_anomaly}, threat detection~\cite{app_threat}, traffic control~\cite{app_traffic} and so on. Image segmentation plays a vital role to segregate objects of interest from the clutter present in natural scenes.

\section{Discussion and Future Scope}
Throughout the paper various methods have been discussed with an effort to highlight their key contributions, pros and cons. With so many different options it is still hard to choose the right approach for a problem. The most optimal way to choose a correct algorithm is to first analyze the variables that affect the choice. 
\par One of the most important aspect that affects the performance of deep learning based approaches is the availability of datasets and annotations. In that regard a concise list of datasets belonging to various domains has been provided in table \ref{tab:dataset}. When working on other small scale datasets it is a common practice to pre-train the network on a larger dataset of a similar domain. Sometimes may be ample amount of samples are available yet pixel level segmentation labels may not be available as creating them is a taxing problem. Even in those cases pre-training parts of networks on other related problems like classification or localization can also help in the process of learning a better set of weights.
\par A related decision that one must take in this regard is to choose among supervised, unsupervised or weakly supervised algorithms. In the current scenario there exists a large number of supervised approaches, however unsupervised and weakly supervised algorithms are still far from reaching a level of saturation. This is a legitimate concern in the field of image segmentation because data collection can be carried out through many automated processes but annotating them perfectly requires manual labor. It is one of the most prominent areas where a researcher can contribute in terms of building end-to-end scalable systems that can model data distribution, decide on the optimal number of classes and create accurate pixel-level segmentation maps in a completely unsupervised domain. Weakly supervised algorithms is also a highly demanding area. It is much easier to collect annotations corresponding to problems like classification or localization. Using those annotations to guide image segmentation problem is also a promising domain.
\par The next important aspect of building deep learning models for image segmentation is the selection of the appropriate approaches. Pre-trained classifiers can be used for various fully convolutional approaches. Most of the time some kind of multi-scale feature fusion can be carried out by combining information from different depths of the network. Pre-trained classifiers like VGGNet or ResNet or DenseNet are also often used for the encoder part of a encoder-decoder architecture. Here also information can be passed from various layers of encoders to corresponding similar sized layers of the decoder to obtain multi-scale information. Another major benefit of encoder-decoder architectures are that if the down-sampling and up-sampling operations are designed carefully, outputs can be generated which are of the same size as that of the input. It is a major benefit over simple convolutional approaches like FCN or DeepMask. This removes the dependency on the input size and hence makes the system more scalable. These two approaches are the most common in case of semantic segmentation problem. However, if finer level of instance specific segments are required it is often necessary to couple with other methods corresponding to object detection. Utilizing bounding box information is one way to address these problems, while other approaches use attention based models or recurrent models to provide output as sequence of segments for each instance of the object. 
\par There can be two aspects to consider while measuring the performance of the system. One is speed and the other is accuracy. Conditional random field is one of the most commonly used post-processing module for refining outputs from other networks. CRFs can be simulated as an RNN to create end-to-end trainable modules to provide very precise segmentation maps. Other refinement strategies include the use of over-segmentation algorithms like superpixels, or using human interactions to guide segmentation algorithms. In terms of gain in speed, networks can be highly compressed using strategies like depth-wise separable convolutions, kernel factorizations, reducing number of spatial convolutions and so on. These tactics can reduce number of parameters to a great extent without reducing the performance too much. Lately, generative adversarial networks have seen a tremendous rise in popularity. However, their use in the field of segmentation is still pretty thin with only a handful of approaches addressing the avenue. Given the success they have gained it certainly has potential to improve existing systems by a great margin.
\par The future of image segmentation largely depends on the quality and quantity of available data. While there is an abundance of unstructured data in the internet, the lack of accurate annotations is a legitimate concern. Especially pixel level annotations can be incredibly difficult to obtain without manual intervention. The most ideal scenario would be to exploit the data distribution itself to analyze and extract meaningful segments that represent concepts rather than content. This is an incredibly challenging task especially if we are working with a huge amount of unstructured data. The key is to map a representation of the data distribution to the intent of the problem statement such that the derived segments are meaningful in some way and contributes to the overall purpose of the system.
\section{Conclusion}
Image segmentation has seen a new rush of deep learning based algorithms. Starting with the evolution of deep learning based algorithms we have thoroughly explained the pros and cons of the various state of the art algorithms associated with image segmentation based on deep learning. The simple explanations allow the reader to grasp the most basic concepts that contribute to the success of deep learning based image segmentation algorithms. The unified representation scheme followed in the diagrams can highlight the similarities and differences of various algorithms. In the future this theoretical survey work can be accompanied by empirical analysis of the discussed methods.

\section*{Acknowledgement}
This work is partially supported by the project order no. SB/S3/EECE/054/2016, dated 25/11/2016, sponsored by SERB (Government of India) and carried out at the Centre for Microprocessor Application for Training Education and Research, CSE Department, Jadavpur University. The authors would also like to thank the reviewers for their valuable suggestions which gelp to improve the quality of the manuscript.

\pagebreak

\section*{Supplementary Information}
\subsection*{Image Segmenation before deep learning : A refresher}
\paragraph*{Thresholding : }
The most straight forward image segmentation can be assigning class labels to each image based on a threshold-point with respect to the intensity values. One of the earliest algorithm popularly known as Otsu's~\cite{met14_otsu} method chooses a threshold point with respect to maximum variance. Many modern approaches have been applied which involving fuzzy logic~\cite{met6_fuzzy} or non-linear thresholds~\cite{met2_hist}. While early approaches were mainly focused on binary thresholding, multi-class segmentation have also come up in subsequent years.
\paragraph*{Clustering methods : }
Clustering methods like K-means~\cite{met15_kmeans} can cluster images into more than one class. It is quite a simple process which can yield excellent results for images where objects of interests are in a high contrast with respect to the background. Other clustering approaches are combined with fuzzy logic ~\cite{bezdek1984fcm,anirban2} or even multi-objective optimizations~\cite{anirban1}.
\paragraph*{Histogram-based methods : }
Histogram-based methods~\cite{met1_hist,met2_hist} provide a more global perspective when it comes to semantic segmentation. By analyzing the peaks and troughs of the histogram the image can be appropriately segmented into an optimal number of segments. Unlike clustering algorithms like k-means number of clusters need not be known beforehand. 
\paragraph*{Edge detection : }
Another angle to look at the problem of image segmentation is to consider the semantic boundaries~\cite{met5_edge} between objects. Semantic image segmentation is quite related to edge detection algorithms for many reasons such as individual objects tend to be separated in an image by an edge that exhibits a sharp change in the intensity gradient. A common method that utilizes concepts of edge-detection and semantic segmentations are super-pixel based processes~\cite{met24_superpixel_das,met25_contour_choudhuri2016multi}.
\paragraph*{Region-growing methods : }
While intensity based methods are quite potent in clustering images, they do not consider the factor of locality. Region growing methods rely on the assumption that neighboring pixels within common segment share some common properties. These kind of methods generally start from seed points and slowly grow while staying within semantic boundaries~\cite{survey9_region} The regions are grown by merging adjacent smaller regions based on some intra-region variance or energy. Many common algorithms such as Mumford-Shah~\cite{met17_mumford} or Snakes algorithm~\cite{met18_snakes} Other variants of such methods depend on the lambda connectedness and grown on the basis of pixel intensities
\paragraph*{Graph based approaches : }
Graph partitioning algorithms can be used to consider context of locality by treating pixels or groups of pixels as nodes thus converting an image to a weighted undirected graphs. Graph cutting algorithms ~\cite{met20_normalized,boykov2001fast,boykov2001interactive,rother2004grabcut} may be efficiently used to obtain the segments. Probabilistic graphical models such as Markov random fields(MRF)~\cite{met21_markov} can be used to create a pixel labeling scheme based on prior probability distribution. MRF tries to maximize the probability of correctly classifying pixels or regions based on a set of features.Probabilistic graphical models like MRF or other similar graph based approaches ~\cite{met19_graph} can also be seen as energy minimization problems~\cite{met7_entropy}. Simulated annealing~\cite{met22_simAnnealing} can be an apt example in this regard as well. These approaches can choose to partition graphs based on their energy.
\paragraph*{Watershed transformations : }
Watershed algorithms~\cite{met23_watershed} assumes the gradients of an image as a topographic surface. Analogous to the water flow lines in such a surface, the pixels with the highest gradients act as contours for segmentation

\paragraph*{Feature based techniques : }
Various features such as colors, textures, shape, gradients, and so on can be used to train machine learning algorithms such as neural networks~\cite{met10_hopfield,met12_sofm,met9_pcnn} or support vector machines to perform pixel level classification. Before the onset of deep learning fully connected neural networks were efficiently used to perform semantic segmentation. However fully connected networks can incur huge memory challenges for larger images as each layer is accompanied by trainable weight in the order of $O(n^2)$, where n is the height and weight of the final activation map of each input image. 

\subsection*{A brief history of neural networks}
With the initial proposition of artificial neurons by McCulloch and Pitts~\cite{hist_mcculloch1943} and the follow up model of perceptron~\cite{hist_rosenblatt1958perceptron}, one could simulate linear models with learnable weights. But the limitation of linear systems soon crept up as they were found unable to learn cases with dependent variables like in case of the XOR problem ~\cite{hist_papert1967linearly}. It took almost a decade before multi-layer models started to arrive since the introduction of the Neocognitron ~\cite{hist_fukushima1982neocognitron, hist_fukushima1988neocognitron}, a hierarchical self-organizing neural architecture. The problem was however to extended the idea of stochastic gradient based learning over multiple layers. That is when the idea of back-propagation ~\cite{hist_backprop_rumelhart1986learning} surfaced. With back-propagation the error from the visible layers could be propagated as a chain of partial derivatives to the weights in the intermediate layers. Introduction of non-linear activation functions such as sigmoid units allowed these intermediate gradients to flow without break as well as solve the XOR problem. While it was clear that deeper networks provided better learning capability but they also induce vanishing or exploding gradients ~\cite{hist_vanishing_bengio1994learning}. This was a difficult problem to deal with specially in sequential networks until long short-term memory cells~\cite{deep_lstm} were proposed to replace tradition recurrent neural networks~\cite{deep_rnn}. Also the convolutional neural networks~\cite{deep_lenet} were introduced as an end-to-end classifier which became one of the most important contribution of the modern computer vision domain. Further down a decade deep learning started flourishing started with Hinton proposing the restricted boltzmann machines~\cite{deep_layerwise_rbm} and the deep belief networks~\cite{deep_dbn}.

\paragraph*{Advancement in Hardwares : }
Another important factor that triggered onset of deep learning was the availability of parallel computation capabilities. Earlier dependence on CPU based architecture was creating a bottleneck for the large amount of floating point operations. Clusters were something that was costly and hence research was quite localized at organizations with substantial funding. But with the onset of graphics processing units(GPUs)~\cite{misc_cpuGPU} developed by Nvidia and their underlying CUDA api for accessing the parallel compute cores neural learning systems got a significant boost. These graphics cards provided hundreds and thousands of computational cores at a much cheaper rate which were efficiently built to handle matrix based operation which is ideal for neural networks.	

\paragraph*{Larger Datasets : }
With the availibilty of GPUs neural network based research became much more widespread. But another important factor to boost it was the influx of new and challenging datasets and challenges. Starting with the MNIST dataset~\cite{data_mnist}, digit recognition became one of the vanilla challenges for new systems. In the late 90's Yann LeCun, Yoshua Bengio and Geoffrey Hinton kept deep learning alive at the Canadian institute of advanced research by launching the CIFAR datasets~\cite{data_cifar} of natural objects for object classification. The challenge was further extended to new heights with the 1000 class Imagenet database~\cite{data_imagenet} along with the competition called Imagenet large scale visual recognition challenge~\cite{data_ILSVRC}. Till this date this has been one of the main challenges which acts as a benchmark for any new  object classification algorithms. As people moved on to more complex datasets, image segmentation became challenging as well. Many datasets and competition such as PASCAL VOC Image segmentation~\cite{data_pascal}, ILSVRC Scene parsing~\cite{data_ILSVRC}, DAVIS Challenge~\cite{data_DAVIS}, Microsoft COCO~\cite{data_COCO} and so on came into the picture that boosted research in image segmentation as well.

\subsection*{Broad categories of typical deep learning models}
\begin{figure}[htbp]
	\centering
	\includegraphics[width=0.4\textwidth]{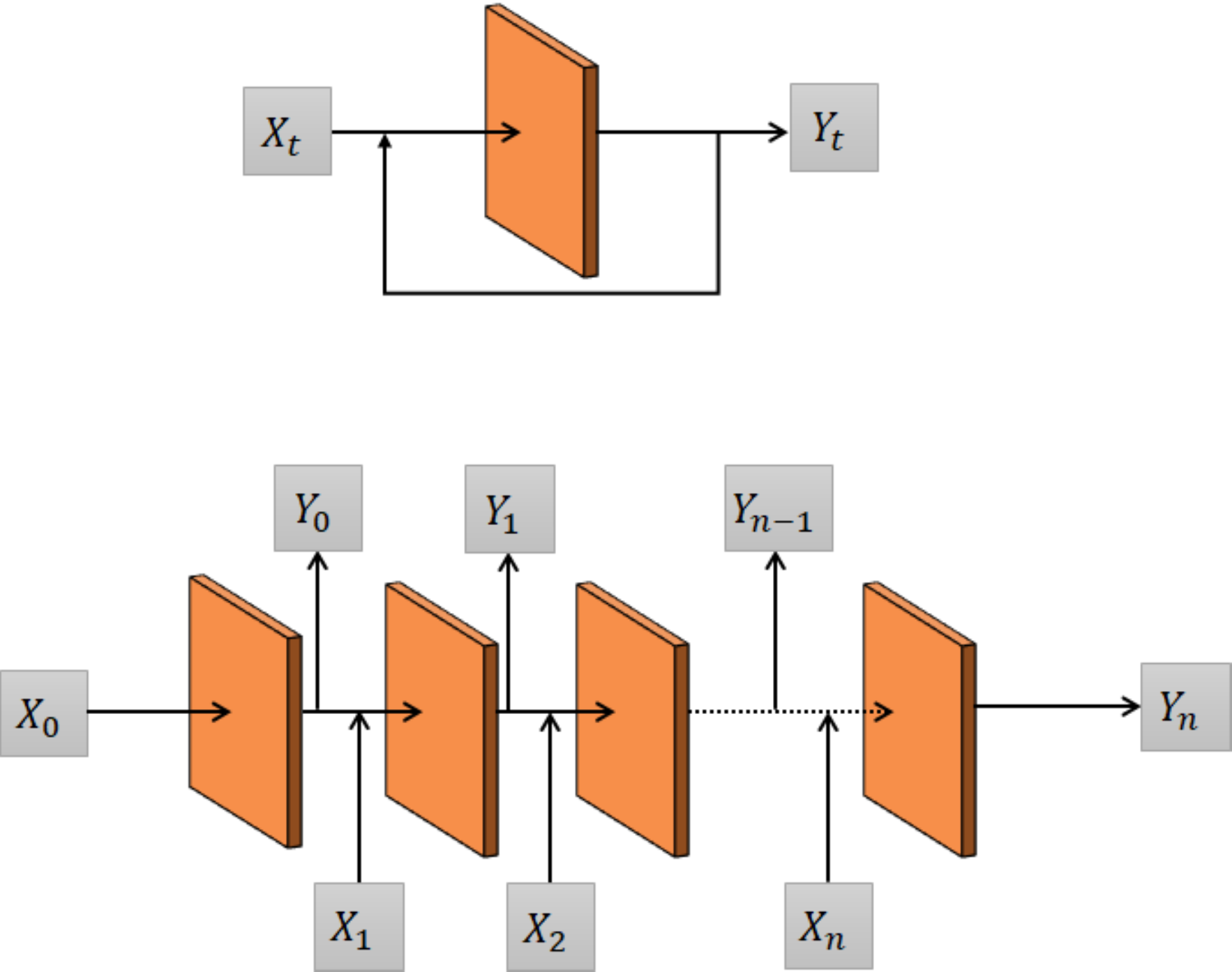}
	\includegraphics[width=0.4\textwidth]{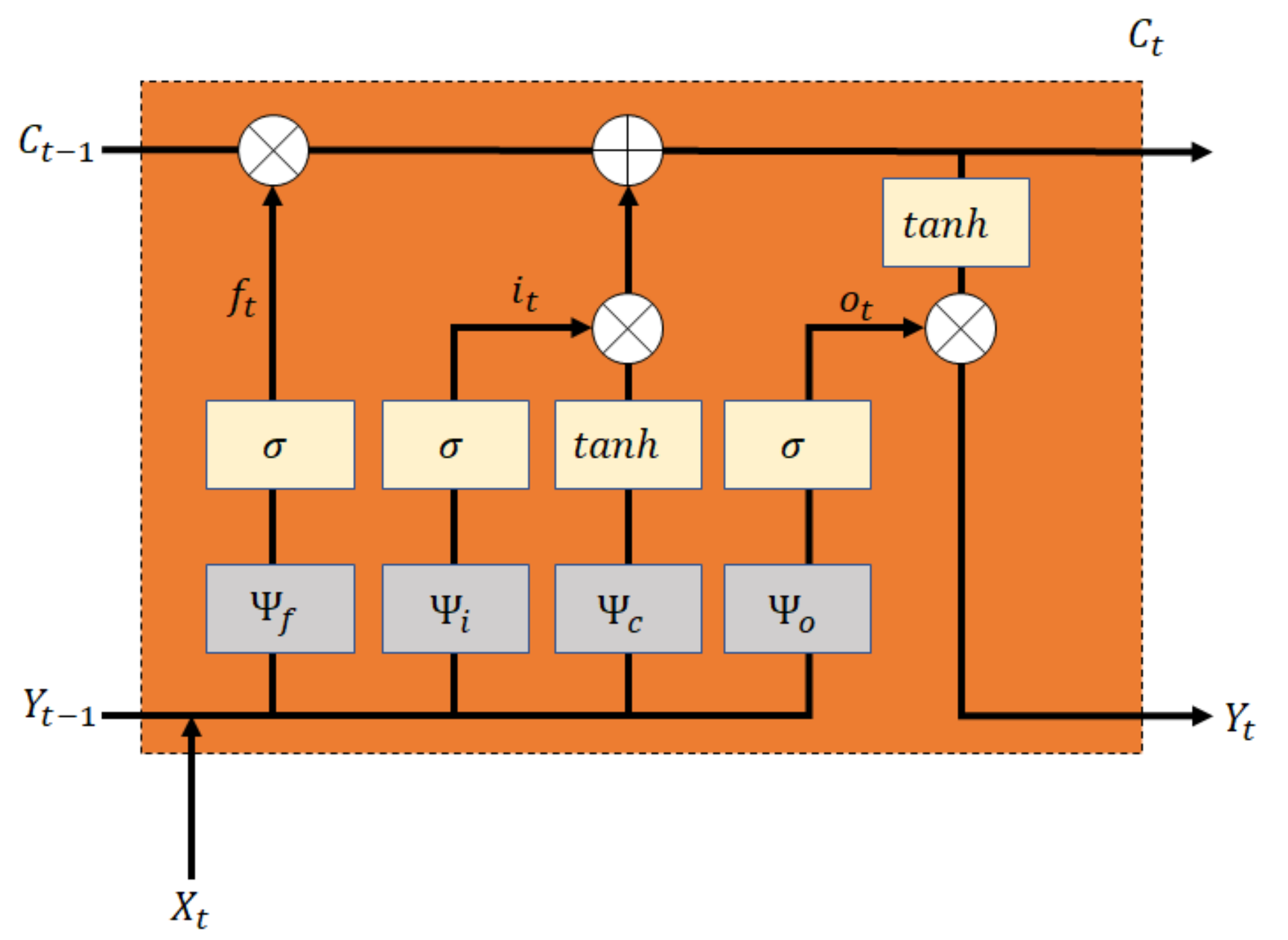}
	\caption{Sequential Models: (topleft) Generic Representation for $t$-th input, (bottomleft) Unfolded network along a sequence of $n$ inputs, (Right) A generic LSTM Module. The $\Psi$ function represents a linear layer in tradition LSTM and a convolutional layer in convolutional LSTM}
	\label{fig:sequential}
\end{figure}

With the advent of deep learning many interesting neural networks were proposed that solved various challenges
\paragraph*{Sequential Models : }
The earliest problem with deep networks were seen with recurrent neural networks~\cite{deep_rnn}. Recurrent networks can be characterized by the feedback loop that allows them to accept a sequence of inputs while carrying the learned information over every time-step. However, over long chains of inputs it was seen that information got lost over time due to vanishing gradients. Vanishing or exploding gradients primarily occurs due to the long multiplication chains of partial derivatives which can push resultant value to almost zero or a huge value, which in turn results in either an insignificant or too large weight update~\cite{hist_vanishing_bengio1994learning}. The first attempt to solve this was proposed by Long short term memory~\cite{deep_lstm} architectures where relevant information can be propagated through long distances through a highway channel which was affected only my addition or subtraction, hence, preserving the gradient values.	Sequential models can have various applications in computer vision such as video processing, instance segmentation and so on. Fig. ~\ref{fig:sequential} shows a sample of a generic and unrolled version of a typical recurrent network.
.

\begin{figure}[htbp]
	\centering
	\includegraphics[width=0.5\textwidth]{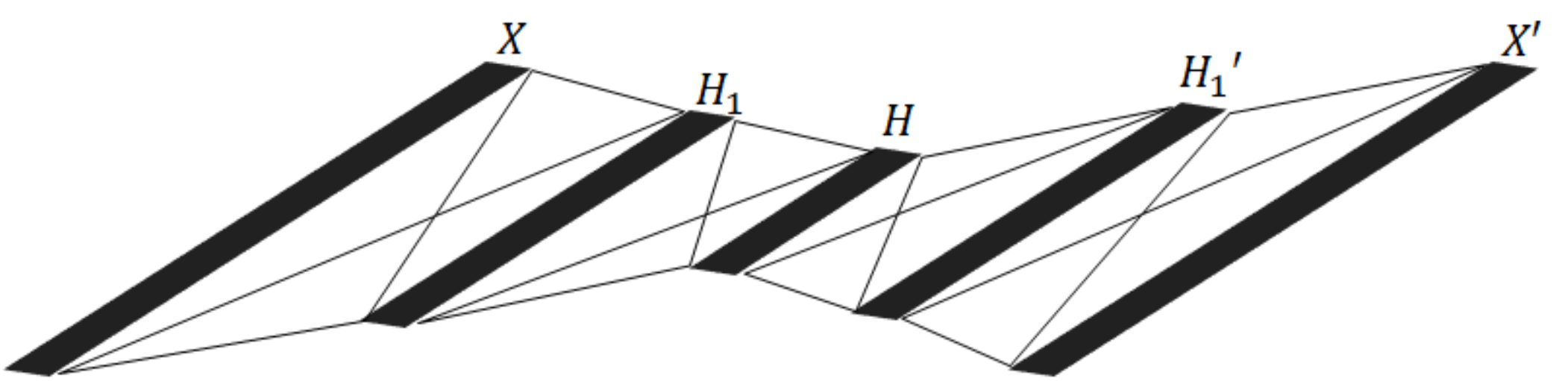}
	\caption{Generic representation of autoencoder with fully connected linear layers}
	\label{fig:autoencoder}
\end{figure}

\paragraph*{Autoencoders : }
Autoencoders have been around since the introduction of auto-associative networks~\cite{deep_autoassociative} for multi-layer perceptron. The principle behind autoencoders is to encode raw inputs into a latent representation. A decoder network attempts to reconstruct the input from the encoded representation. The minimization of a loss function based on the difference of the input and the reconstructed output ensures minimum loss of information in the intermediate representation. This hidden representation is a compressed form of the actual input. As it preserves most of the defining properties of the input image it is often used as features for further processing.  An autoencoder consists of two main phases, namely, encoding and decoding phase. After training the encoder can be easily used as a feature extractor. The decoder part can be used for generative purposes. Many works use the generative property of the decoder for various image segmentation application~\cite{deep_vae}. The figure below(~\ref{fig:autoencoder}) shows a representation of an autoencoder with fully connected linear layers. 

\begin{figure}[htbp]
	\centering
	\includegraphics[width=0.75\textwidth]{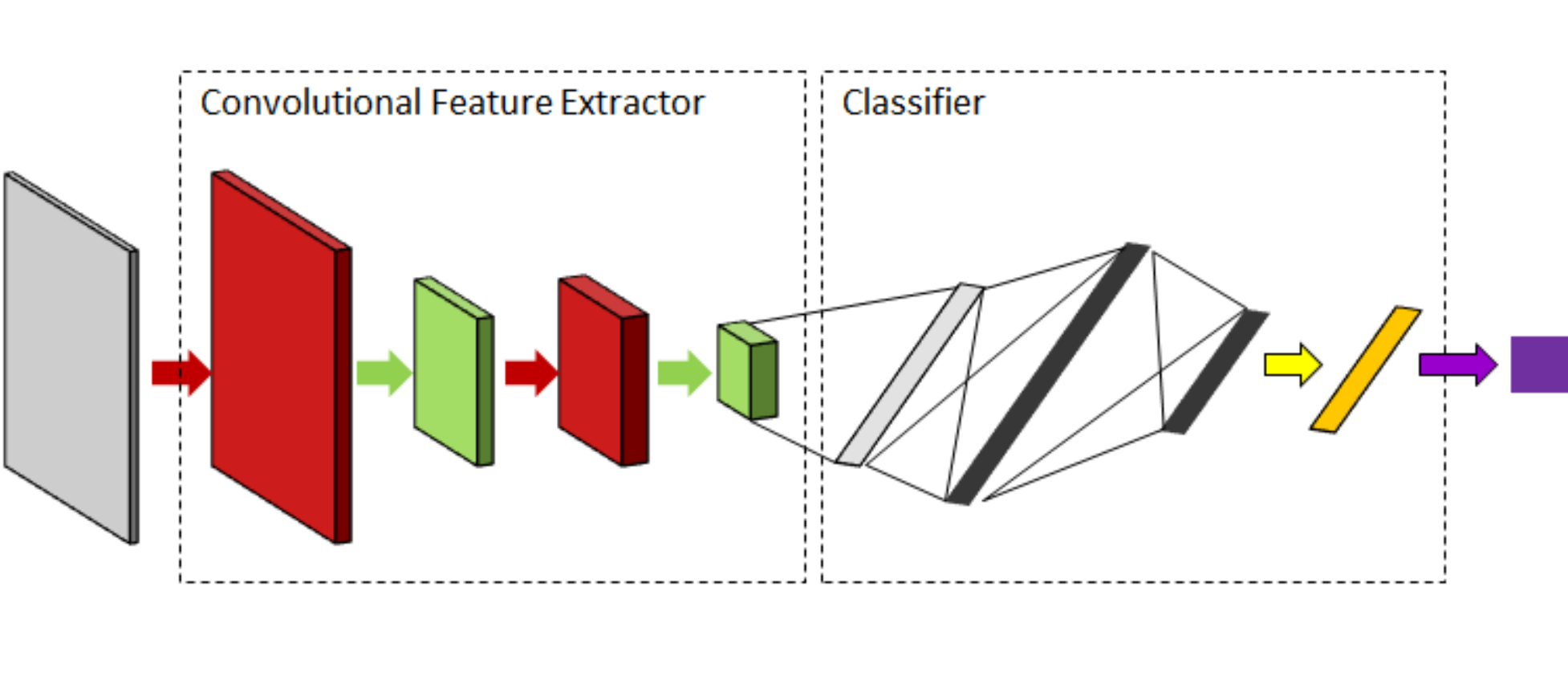}
	\caption{A typical convolutional neural network}
	\label{fig:cnn}
\end{figure}

\begin{figure}[htbp]
	\centering
	\includegraphics[width=0.5\textwidth]{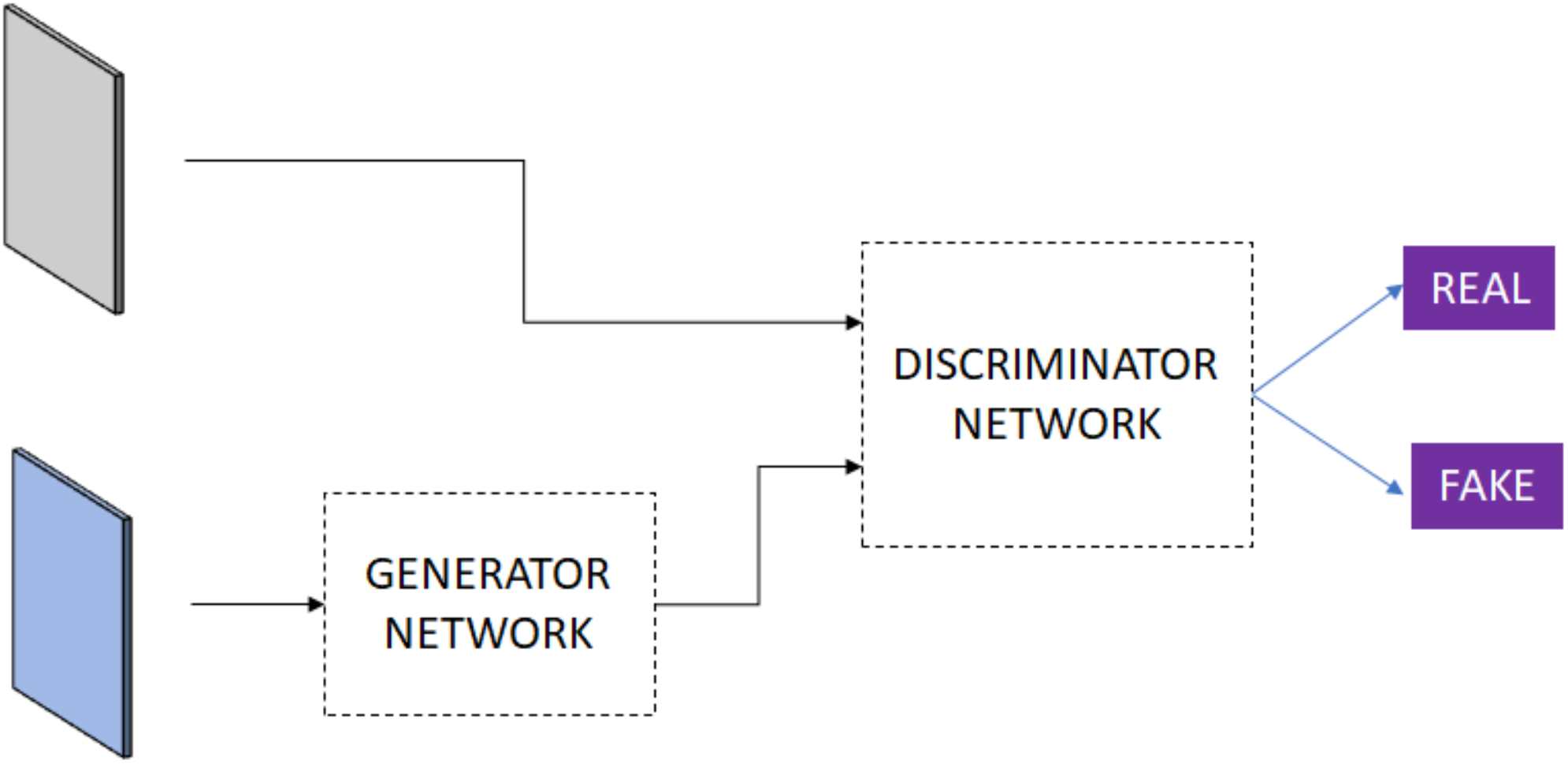}
	\caption{A block diagram of generative adversarial network}
	\label{fig:gan}
\end{figure}

\paragraph*{Convolutional Neural Networks : }
Convolutional neural networks~\cite{deep_lenet,deep_alexnet} are probably one of the most significant inventions under the wing of deep learning for computer vision. Convolutional kernels have been often used for feature extraction from complex images. However designing kernels was not an easy task especially for complex data like natural images. With convolutional neural networks kernels can be randomly initialized and updated iteratively through back propagation based on a error function like cross entropy or mean square error. Many other operations are commonly found in CNNs such as pooling, batch normalization, activations, residual connections and so on. Pooling layer increase the receptive fields of convolutional kernels. Batch normalization~\cite{deep_batchnorm} refers to a generalization process that involves normalization of activations across the batch. Activation functions have been an integral part of perceptron based learning. Since the introduction of AlexNet~\cite{deep_alexnet}, rectified linear units (ReLU)~\cite{deep_relu} have been the activation function of choice. ReLU(Rectified Linear Unit) provides a gradient of either 0 or 1, thus, preventing vanishing or exploding gradients and also inducing sparsity in the activations. Lately another interesting method for gradient boosting was seen in the application residual connection. Residual connections~\cite{deep_resnet} provided an alternate path for gradients to flow which is devoid of operations that inhibit gradients. Residual connections have also been applied in many cases to improve the quality of segmented images. Fig.~\ref{fig:cnn} shows a convolutional feature extractor along with fully connected classifier.
\paragraph*{Generative Models : }
Generative models are probably one of the latest attractions of deep learning in computer vision. While sequential models like long short term memory or gated recurrent units are able to generate sequence of vectorized elements, in computer vision it is much more difficult due to the spatial complexities. Lately various methodologies like variational autoencoders~\cite{deep_vae}, or adversarial learning~\cite{deep_aae,deep_gan} has become extremely efficient in generating complex images. The generative properties can be used quite efficiently in tasks like generation of segmentation masks. An typical example of a generative network is shown in fig.~\ref{fig:gan} which learns by adversarial learning.
\bibliographystyle{ieeetr}
\bibliography{ref}

\end{document}